\documentclass[10pt,twocolumn,letterpaper]{article}

\usepackage{iccv}

\usepackage{graphicx}
\usepackage{amsmath}
\usepackage{amssymb}
\usepackage{booktabs}
\usepackage{kotex}
\usepackage{times}
\usepackage{epsfig}
\usepackage{mathtools}
\usepackage{dsfont}
\usepackage{cuted} 
\usepackage[accsupp]{axessibility}

\usepackage{multirow}
\usepackage{subcaption}
\usepackage[noend]{algpseudocode}
\usepackage{float}
  \floatstyle{ruled}
  \newfloat{algorithm}{thp}{alg}
\floatname{algorithm}{Algorithm}
\usepackage{algorithmicx}
\usepackage[dvipsnames]{xcolor}
\definecolor{SigmaColor}{rgb}{0.98,0.45,0.0}

\newcommand\sbullet[1][.5]{\mathbin{\vcenter{\hbox{\scalebox{#1}{$\bullet$}}}}}

\iccvfinalcopy 

\def\iccvPaperID{6139} 

\usepackage[pagebackref=true,breaklinks=true,letterpaper=true,colorlinks,bookmarks=false]{hyperref}

\usepackage[capitalize]{cleveref}
\crefname{section}{Sec.}{Secs.}
\Crefname{section}{Section}{Sections}
\Crefname{table}{Table}{Tables}
\crefname{table}{Tab.}{Tabs.}

\begin{document}

\title{Locomotion-Action-Manipulation: \\Synthesizing Human-Scene Interactions in Complex 3D Environments}

\author{Jiye Lee\\
Seoul National University\\
{\tt\small kay2353@snu.ac.kr}
\and
Hanbyul Joo\\
Seoul National University\\
{\tt\small hbjoo@snu.ac.kr}
}
\maketitle 

\begin{strip}\centering
\includegraphics[width=\linewidth, trim={0 0cm 0 0.7cm},clip]{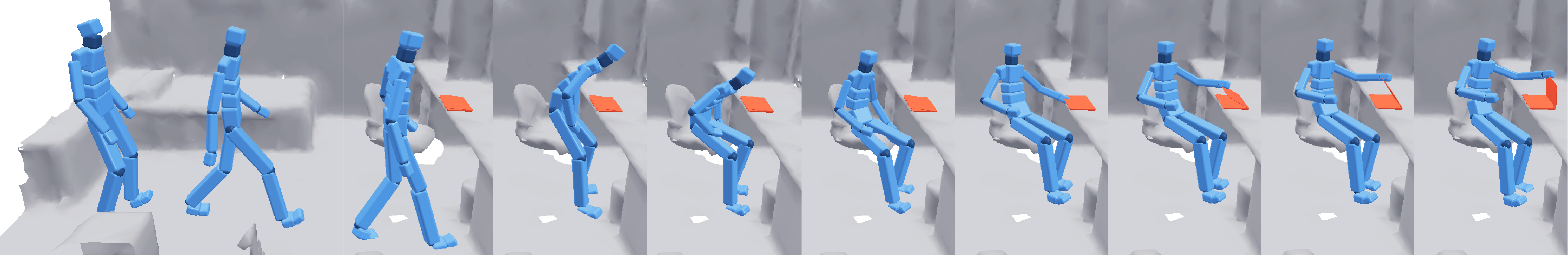}
\captionof{figure}{Our system, LAMA, produces high-quality and realistic 3D human motions that include locomotion, scene interactions, and manipulations within a given 3D scene and designated interaction cues. 
\label{fig:teaser}}
\end{strip}

\begin{abstract}
Synthesizing interaction-involved human motions has been challenging due to the high complexity of 3D environments and the diversity of possible human behaviors within.
We present LAMA, \textbf{L}ocomotion-\textbf{A}ction-\textbf{MA}nipulation, to synthesize natural and plausible long- term human movements in complex indoor environments. The key motivation of LAMA is to build a unified framework to encompass a series of everyday motions including locomotion, scene interaction, and object manipulation. Unlike existing methods that require motion data ``paired'' with scanned 3D scenes for supervision, we formulate the problem as a test-time optimization by using human motion capture data only for synthesis. LAMA leverages a reinforcement learning framework coupled with a motion matching algorithm for optimization, and further exploits a motion editing framework via manifold learning to cover possible variations in interaction and manipulation.
Throughout extensive experiments, we demonstrate that LAMA outperforms previous approaches in synthesizing realistic motions in various challenging scenarios.
Project page: \href{https://jiyewise.github.io/projects/LAMA/}{https://jiyewise.github.io/projects/LAMA/}.

\end{abstract}
\section{Introduction}
\label{sec:intro}

Synthesizing interactions within real-life 3D environments has been a challenging research problem due to its complexity and diversity. 
The spatial constraint arising from real-life 3D environments where many objects are cluttered makes motion synthesis highly constrained and complex. 
Furthermore, the nearly indefinite diversity of possible spatial arrangements of the 3D environment and human interaction behaviors 
makes generalization in synthesis difficult. 
%

Due to the wide range of technical challenges involved in human-scene interactions, previous approaches have focused on sub-problems, such as (1) modeling static poses~\cite{kim2014shape2pose,zhang2020place,zhang2020generating,hassan2021posa,xie2022chore,savva2016pigraphs,Zhao:ECCV:2022} or (2) human object interactions with a single target object or interaction type~\cite{starke2019neural,zhang2022couch, zhang2021manipnet,taheri2022goal,taheri2020grab,qin2022dexmv,yang2022chopsticks,eom2019mpc}. More recent methods~\cite{wang2021synthesizing,wang2022towards,hassan2021samp} extend to synthesizing dynamic interaction motions in real-world 3D scenes, where they use ``scene-paired'' motion datasets~\cite{hassan2019prox}  in which motion is simultaneously captured with the surrounding 3D environment.
As such paired dataset is rare and difficult to scale up, the performance of these methods is fundamentally limited in fully covering the complexity and diversity of human interaction in real-world 3D scenes.


In this paper, we present LAMA, \textbf{L}ocomotion-\textbf{A}ction-\textbf{MA}nipulation, to synthesize natural and plausible long-term human motions in complex indoor environments. 
The key motivation of LAMA is to build a unified framework covering a series of everyday motions within real-world 3D scenes: locomotion through cluttered areas, interaction with the scene, and manipulation of objects. 
Unlike previous approaches~\cite{wang2021synthesizing, hassan2021samp} that use a ``scene-paired'' motion dataset  for supervision, we formulate it as a test-time optimization by utilizing only human motion capture data.
Exploiting reinforcement learning (RL) as a tool for optimization,
we present an RL-based framework coupled with a motion matching algorithm~\cite{clavet2016motion, buttner2015motion} to synthesize locomotion and scene interaction seamlessly while adapting to complex 3D scenes with collision avoidance handling. The object manipulation in our framework is performed via a motion editing approach on top, by learning an autoencoder-based motion manifold space~\cite{holden2016deepsynthesis}. 
As a test-time optimization framework, LAMA is applicable to any 3D scene scenarios (e.g., public datasets or any newly scanned scenes).
Through extensive quantitative and qualitative evaluations against existing methods, we demonstrate that our method outperforms~\cite{wang2021synthesizing, hassan2021samp} in various challenging scenarios. 

Our contributions are summarized as follows:  (1) The first method to generate realistic long-term motions combined with locomotion, scene interaction, and manipulation in complex 3D scenes without ``paired" datasets;
(2) A novel test-time optimization framework requiring human motion capture data only by incorporating a reinforcement learning framework coupled with motion matching, equipped with well-designed state and rewards for collision avoidance and scene interactions;
(3) the state-of-the-art motion synthesis quality with longer duration (near 10 sec);
(4) A newly captured and polished motion capture dataset including locomotion and action (e.g., sitting) suitable for motion matching.

\section{Related Work}

\textbf{Generating Human-Scene Interactions.}
Generating natural human motion has been a widely researched topic in the computer vision community. Early methods focus on synthesizing or predicting human movements by exploiting neural networks~\cite{habibie2017recurrent,fragkiadaki2015recurrent,li2021aict,li2021aict, martinez2017human,taylor2009factored,villegas2017longterm,petrovich2021action}. However, these approaches primarily address the synthesis of human motion itself, without taking into account the surrounding 3D environments. Recent approaches begin to tackle modeling and synthesizing human interactions within 3D scenes, or with objects. Many focus on statically posing humans within the given 3D environment~\cite{kim2014shape2pose, zhang2020place,zhang2020generating,hassan2019prox}, by generating human scene interaction poses from various types of input including object semantics~\cite{hassan2021posa}, images~\cite{huang2022behave,zhang2020phosa,xu2021d3d,xie2022chore,jiang2022neuralhofusion,jiang2022chairs}, and text descriptions~\cite{savva2016pigraphs,Zhao:ECCV:2022}. 


Recently, there have been approaches to synthesize dynamic human-object interactions (e.g., sitting on chairs, carrying boxes). 
Starke et al.~\cite{starke2019neural} introduce an autoregressive learning framework with object geometry based environmental encodings to synthesize human-object interactions. Although the encoding includes information on multiple objects within a scene, as demonstrated in~\cite{hassan2021samp}, no explicit module for navigating through cluttered 3D scenes exists in~\cite{starke2019neural}. 
Later work~\cite{hassan2021samp,zhang2022couch} extends this by synthesizing motions conditioned with variations of objects and contact points. Other approaches~\cite{zhang2021manipnet,taheri2022goal, wu2022saga, taheri2020grab,qin2022dexmv,yang2022chopsticks, zhou2022toch} focus on generating natural hand movements for manipulation, which is extended by including full body motions~\cite{taheri2022goal,wu2022saga}.
Physics-based character control to synthesize human-object interactions has been also explored in \cite{merel2020catch,eom2019mpc,chao2021chair,qin2022dexmv,yang2022chopsticks, lee2022comcon}. Although these methods cover a variety of human-object interactions, most of them focus on a specific interaction type or the relationship between the human and the target object without long-term navigation in cluttered 3D scenes. 

\begin{figure}[t]
\includegraphics[width=\linewidth, trim={0 0 0 0},clip]{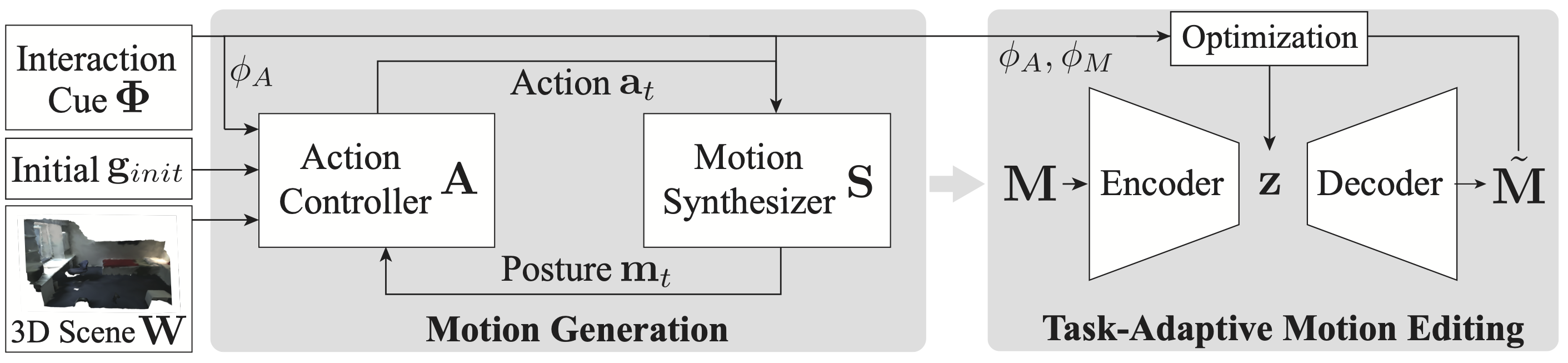}
\caption{Overview of LAMA.}
\label{fig:overview}
\vspace{-.2in}
\end{figure}

More recent approaches include generating natural human scene interactions in cluttered 3D scenes~\cite{wang2021synthesizing,wang2022towards,cao2020long,wang2020motion}, closely related to ours. 
These methods are trained using human motion datasets paired with 3D scenes, which require both ground truth motion and simultaneously captured 3D scenes for supervision. 
Due to difficulties in acquiring such data, some methods exploit synthetic datasets~\cite{cao2020long,wang2020motion}, data fitted from depth videos~\cite{wang2021synthesizing}, or motion snapshots with short duration (1-3 sec)~\cite{wang2022humanise}.
In previous approaches~\cite{hassan2021samp,wang2022towards},  navigation in cluttered environments is often performed by a separate module via path planning (e.g., $A^*$ algorithm) by approximating the volume of a human as a cylinder. These path planning based methods approximate the spatial information of the scene and the body and therefore have limitations under highly complex conditions. 

\textbf{Motion Synthesis and Editing.}
Synthesizing natural human motions by leveraging motion capture data has also been a long-researched topic in computer graphics. Some approaches~\cite{lee2002interactive,lucas2002motion} construct motion graphs, where plausible transitions are inserted as edges, and motion synthesis is done by traversing through the graph. Similar approaches~\cite{lee2006motion,shum2008interaction} connect motion patches to synthesize interactions in a virtual environment or multi-person interactions. Due to its versatility and simplicity, variations have been made to the graph-based approach, such as motion grammar~\cite{hyun2016motion} which enforces traversing rules in the motion graph. Motion matching~\cite{buttner2015motion,clavet2016motion} can also be understood as a special case of motion graph traversal, where the plausible transitions are not precomputed but searched during runtime.
%
Recent advances in deep learning allow to leverage motion capture data for motion manifold learning~\cite{holden2016deepsynthesis,starke2022deepphase,holden2017phase}. Autoregressive approaches based on variational autoencoders (VAE)~\cite{ling2020motionvae,petrovich2021action} and recurrent neural networks~\cite{lee2018interactive,harvey2020robust,park2019icc} are also used to forecast future motions based on past frames.
These frameworks are generalized to synthesize a diverse set of motions including locomotion on terrains~\cite{holden2017phase} mazes~\cite{ling2020motionvae}, action-specified motions~\cite{petrovich2021action}, and interaction-involved sports~\cite{lee2018interactive,park2019icc}. 
Neural network-based methods are also reported to be successful in various motion editing tasks such as skeleton retargeting~\cite{aberman2020skeleton}, style transfer~\cite{holden2016deepsynthesis,aberman2020unpaired}, and in-betweening~\cite{harvey2020robust}.
	
 Reinforcement learning (RL) has also been successful in combination with both data-driven and physics-based approaches for synthesizing human motions. Combined with data-driven approaches, RL serves as a control module that generates corresponding motions to a given user input by traversing motion graphs~\cite{lee2004precomputing}, latent space~\cite{levine2012continuous,treuille2007near,ling2020motionvae}, and precomputed transition tables~\cite{lee2021timecritical}. Deep reinforcement learning (DRL) has been widely used as well to synthesize physically plausible movements with a diverse set of motor skills~\cite{peng2018deepmimic,park2019icc,lee2021parametric,bergamin2019drecon,won2020scalable,peng2021amp,peng2022ase, lee2022comcon}. The key idea of these methods comes from imitation learning, where the control policy in DRL is optimized to actuate the character based on the character's physical state, to meet the goal of tracking the given reference motion in a physically simulated environment.

\section{Method}\label{sec:method}


\subsection{Overview}\label{sec:notations}
Our system, dubbed as $LAMA$, outputs a sequence of human poses $\mathbf{M}=\{\mathbf{m}_t\}_{t=1}^T$ by taking the 3D scene $\mathbf{W}$, desired interaction cues $\mathbf{\Phi}$, and initial state $\mathbf{g}_{init} = (\mathbf{p}^0_{root}, \mathbf{r}^0_{root})$ as inputs:
\begin{equation}
    \mathbf{M} = \mathcal{LAMA}(\mathbf{W}, \mathbf{\Phi}, \mathbf{g}_{init}  ),
\end{equation}
where $\mathbf{p}^0_{root} \in \mathbb{R}^3$ and $\mathbf{r}^0_{root} \in so(3)$  represent the global position and orientation of the character's root respectively at initial (i.e., $t=0$). 
The output posture at time $t$, $\mathbf{m}_t =(\mathbf{p}^t_{root}, \mathbf{r}^t_{root}, \mathbf{r}^t_1,..., \mathbf{r}^t_J)  \in \mathbb{R}^{3J+6}$, is represented by a concatenated vector of global position and orientation of the root, and the local joint orientations of $J$ joints where each $j$-th joint is in angle-axis representations $\mathbf{r}^t_j \in so(3)$. 
Throughout our system, the skeleton tree structure and joint offsets are fixed as shown in Fig.~\ref{fig:skel} (a).
We represent the 3D scene  $\mathbf{W} = \{\mathbf{w}_i\}_{}$  as a set of 3D object and scene meshes, including the background scene mesh and other object meshes targeted for manipulation.

The interaction cues, $\mathbf{\Phi} = [\phi_A, \phi_M]$, represent the expected goal that the output needs to fulfill, and consist of the action cue $\phi_{A}$ for the action task (e.g., sitting) and the manipulation cue $\phi_{M}$ for the manipulation task.
The action cue $\phi_{A}= \{ \mathbf{q}_{root}, \mathbf{r}_{root}, \mathbf{q}_{rFoot}, \mathbf{q}_{lFoot},  \}$ indicates desired position and orientation of the root, and the positions of the left foot and right foot respectively (i.e., $\mathbf{q}_j \in \mathbb{R}^3$). The foot positions are optional and can be automatically determined if not provided. $\phi_{A}$ can be manually chosen to instruct the character or can be automatically given via an off-the-shelf estimator such as GoalNet in~\cite{hassan2021samp}. 
The manipulation cue $\phi_{M}= \{ \mathbf{q}^t_j \}_{j \in J_{M}}$ indicates desired locations of selected joints $J_{M}$ for control, which is mainly used in the motion editing procedure (Sec.~\ref{sec:editing}). For example, $\phi_{M}$ can specify the hand joint trajectory to perform the opening laptop motion in Fig.~\ref{fig:teaser} and Fig.~\ref{fig:ic}. Examples of the 3D scene $\mathbf{W}$, action cue $\mathbf{\phi}_A$ and manipulation cue $\phi_M$ are in Fig.~\ref{fig:ic} (left).  


LAMA is designed via a three-level system composed of action controller $\mathbf{A}$, motion synthesizer $\mathbf{S}$, followed by a manifold-based motion editor $\mathbf{E}$.
The locomotion and action parts are seamlessly performed via the action controller $\mathbf{A}$ and synthesizer $\mathbf{S}$.  The essential idea in our design is to combine the RL framework with motion matching~\cite{clavet2016motion, buttner2015motion} to synthesize realistic human motions while fulfilling the desired scene interaction tasks. 
By taking 3D scene $\mathbf{W}$, action cue $\phi_A$, and initial state $\mathbf{g}_{init}$ as input, the action controller $\mathbf{A}$ makes the use of RL as a way of test-time optimization to synthesize corresponding motion.
A control policy $\pi$ is optimized~\footnote{We use the term ``optimized" rather than ``learned" for the policy since we perform a test-time optimization.} to sample an action at time $t$, 
$\pi(\mathbf{a}_t|\mathbf{s}_t, \mathbf{W}, \phi_A)$, where $\mathbf{a}_t$ indicates the plausible next action containing predicted action types and short-term future forecasting. $\mathbf{s}_t$ is the state cue to represent the current status of the human character including its body posture, surrounding scene occupancy, and the targeting action cue.
Intuitively, action controller $\mathbf{A}$ is optimized to generate plausible next action $\mathbf{a}_t$ by considering the current character-scene state $\mathbf{s}_t$.   
The generated action signal $\mathbf{a}_t$ from the action controller $\mathbf{A}$ is provided as input to the motion synthesizer $\mathbf{S}$, which then determines the posture at the next time step $\mathbf{m}_{t+1}$, i.\,e.,  $\mathbf{S}(\mathbf{m}_t, \mathbf{a}_t) =\mathbf{m}_{t+1}$. The character's next state $\mathbf{s}_{t+1}$ can be computed again from $\mathbf{m}_{t+1}$, which is subsequently taken by the action controller $\mathbf{A}$ as an input for the next time frame.

The initial output motion $\mathbf{M}$ synthesized by $\mathbf{A}$ and $\mathbf{S}$ is followed by a motion editor $\mathbf{E}(\mathbf{M}) = \Tilde{\mathbf{M}}$, where $\Tilde{\mathbf{M}}=\{\Tilde{\mathbf{M}}_t\}_{t=1}^T$ is the edited motion. The goal of the editing module $\mathbf{E}$ is to (1) post-process $\mathbf{M}$ to fit into diverse objects targeted for action task $\phi_A$ (e.g., sitting on a chair with different heights), and (2) perform human-object manipulation instructed by $\phi_M$ (e.g. moving objects, opening doors).
Fig.~\ref{fig:overview} shows the overview of LAMA. 


\subsection{Scene-Aware Action Controller} \label{sec:ac}

Unlike previous approaches~\cite{wang2020motion, hassan2021samp} that use path planning for navigation and a learning-based module trained with a scene-paired motion dataset for interaction, our action controller $\mathbf{A}$ performs locomotion and desired actions seamlessly by fulfilling the action cue $\phi_A$ and avoiding collisions in the 3D scene $\mathbf{W}$. Importantly, given the scene $\mathbf{W}$, action $\phi_A$ and initial state $\mathbf{g}_{init}$ as inputs, our action controller is directly optimized to choose the most plausible motion clip in our motion database at each state, synthesizing natural human motions while taking the 3D scene into account without any scene-paired motion dataset or training procedure. Intuitively, given the current state $\mathbf{s}_t$, the goal of the action controller is to output the best next action $\mathbf{a}_t$ which is used to search the next motion clip in Motion Synthesizer $\mathbf{S}$.

\textbf{State.} The state $\mathbf{s}_t = \psi(\mathbf{m}_{t-1}, \mathbf{m}_{t}, \mathbf{W}, \phi_A )$ at time $t$ is a feature vector representing the current status of the human character, where $\psi$ is the function to compute the state from other inputs. $\mathbf{s}_t = (\mathbf{s}_t^{body}, \mathbf{s}_t^{scene}, \mathbf{s}_t^{inter})$ is composed of body configuration $\mathbf{s}^{body}$, scene occupancy $\mathbf{s}^{scene}$, and desired current target interaction $\mathbf{s}^{inter}$.
Body configuration $\mathbf{s}^{body}= \{ \mathbf{r}, \dot{\mathbf{r}}, \theta_{up}, h, \mathbf{p}_{e} \}$, where $\mathbf{r}, \dot{\mathbf{r}} \in \mathbb{R}^{J\times6}$ are the joint rotations and velocities respectively for the $J$ joints excluding the root in 6D representations~\cite{zhou20196d}, $\theta_{up} \in \mathbb{R}$ is the up vector of the root (represented by the angle w.r.t the Y-axis), $h \in \mathbb{R}$ is the root height from the floor, and $\mathbf{p}_{e} \in \mathbb{R}^{e \times 3}$ is the end-effector position in person-centric coordinate (where $e$ is the number of end-effectors).
$\mathbf{s}^{scene} = \{ \mathbf{g}_{occ}, \mathbf{g}_{root} \}$ includes scene occupancy information in the floor plane, as shown in Fig.~\ref{fig:scenegrid}. $\mathbf{g}_{occ} \in \mathbb{R}^{n^2}$ represents the occupancy grid on the floor plane of neighboring $n$ cells around the agent and  $\mathbf{g}_{root} \in \mathbb{R}^2$ denotes the current global root position of the character in the discretized grid plane. Note that, while we consider the 2D floor grid for efficiency rather than 3D, the 3D scene is still considered via our collision reward term in Sec.~\ref{sec::learnac}.
$\mathbf{s}^{inter}$ represents the action cue the character is targeting, that is $\mathbf{s}^{inter} = \mathbf{\phi}_A$.

\textbf{Action.} Given the current status of the character $\mathbf{s}_t$, the control policy $\pi$ outputs the feasible action $\mathbf{a}_t = (\mathbf{a}_t^{type}, \mathbf{a}_t^{future}, \mathbf{a}_t^{offset})$. $\mathbf{a}_t^{type}$ provides the probabilities of the next action type among possible actions (e.g., walk, sit, or stop), determining the transition timing between actions (e.g., from locomotion to sitting). $\mathbf{a}_t^{future}$ predicts future motion cues such as plausible root position for the next 10, 20, and 30 frames.
Posture offset $\mathbf{a}^{offset}_t$ is intended to modify the raw motion data searched from the motion database in motion synthesizer module $\mathbf{S}$. 
Intuitively, our optimized control policy generates a posture offset $\mathbf{a}^{offset}_t$ to alter the closest plausible raw posture chosen in the database. 
This enables the character to perform more plausible scene-aware human poses only with human motion data.
More details are addressed in Sec.~\ref{sec::synthesizer}.

\begin{figure}[t]
\includegraphics[width=0.9\linewidth, trim={0 0.3 0 0.1},clip]{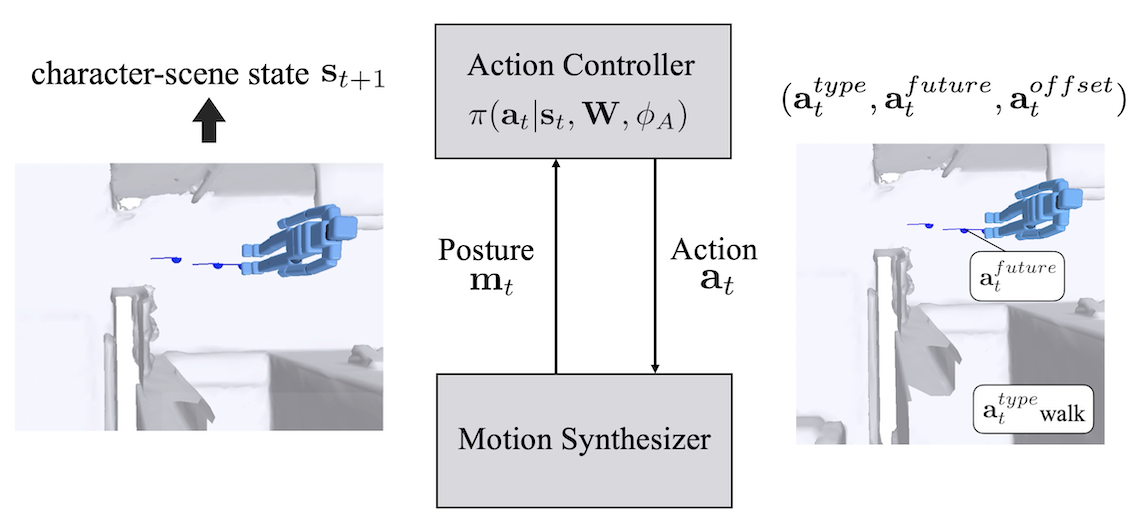}
\caption{Visualization of the relationship between the Action Controller and the Motion Synthesizer.}
\label{fig:acms_coupling}
\vspace{-.2in}
\end{figure}

\subsection{Motion Synthesizer} \label{sec::synthesizer}
By taking the current posture $\mathbf{m}_t$ and actions signal $\mathbf{a}_{t}$ from the action controller $\mathbf{A}$ as inputs, the motion synthesizer $\mathbf{S}$ produces the next plausible posture: $\mathbf{S}(\mathbf{m}_t, \mathbf{a}_t) =\mathbf{m}_{t+1}$. As the first step, the motion synthesizer searches for motion from a motion database that best matches the closest motion feature, then modifies the searched raw motion to be more suitable to the scene. To this end, the motion synthesizer's output $\mathbf{m}_{t+1}$ is in turn fed into the action controller recursively.
We exploit a modified version of the motion matching algorithm~\cite{buttner2015motion, clavet2016motion, holden2020learned} for the first step of motion synthesis. In motion matching, motion synthesis is performed periodically by searching the most plausible next shot motion segments from a motion database, and compositing them into a long connected sequence. 

\textbf{Motion features.} 
Motion feature $\mathbf{y}_t$ represents the characteristic of each frame in the short motion segment and is computed as $ f( \mathbf{m}) = \mathbf{y}_t = \{ \{ \mathbf{p}_\mathrm{j} \},  \{ \dot{\mathbf{p}}_\mathrm{j} \}, \theta_{up}, c, \mathbf{o}_\mathrm{future} \}$.
From a posture $\mathbf{m}$,  the positions and velocities $\mathbf{p}_\mathrm{j}, \dot{\mathbf{p}}_\mathrm{j} \in \mathbb{R}^{3}$ are extracted for the selected joints $j \in \{\mathrm{Head},\allowbreak \mathrm{Hand},\allowbreak \mathrm{Foot}\}$, which are defined in a person-centric coordinate of $\mathbf{m}$. $\theta_{up} \in \mathbb{R}^3$ is the up-vector of the root joint, and $c \in \{0, 0.5, 1\}$ indicates automatically computed foot contact cues of the left and right foot ($0$ for non-contact, $1$ for contact, $0.5$ for non-contact but close to the floor within a threshold). $\mathbf{o}_\mathrm{future} = \{ \{\mathbf{p}_{root}^{\Delta t} \}, \{\mathbf{r}_{root}^{\Delta t} \} \}$ contains the cues for short-term future postures,  where $\mathbf{p}_{root}^{\Delta t}$ and $\mathbf{r}_{root}^{\Delta t}$ are the position and orientation of root joint at $\Delta t$ frames later from the current target frame. $\mathbf{o}_\mathrm{future}$ are computed in 2D XZ plane in person-centric coordinate of the current target motion $\mathbf{m}$, and thus $\mathbf{p}_{root}^{\Delta t}, \mathbf{r}_{root}^{\Delta t} \in \mathbb{R}^2$. The selected future frames are action-type specific, and for locomotion, we extract 10, 20, and 30 frames in the future (at 30Hz) following \cite{clavet2016motion}. 
Intuitively, the motion feature extracts the target frame's posture and temporal cues by considering neighboring frames.
For efficiency, we pre-compute motion features $\mathbf{y}_t$ for every frame of the motion clip in the database. 

Motion feature $\mathbf{x}_t$ of the current state of the character, or the query feature denoted, is also computed in the same way based on posture $\mathbf{m}_{t-1}$, $\mathbf{m}_t$ and $\mathbf{a}_t^{future}$ produced by the action controller, that is $\mathbf{x}_t = f(\mathbf{m}_{t-1}, \mathbf{m}_{t}, \mathbf{a}_t^{type}, \mathbf{a}_t^{future})$. The component  $\mathbf{a}_t^{future}$ serves as $\mathbf{o}_\mathrm{future}$ in the query feature, which can be understood as the action controller providing cues for predicted future postures.

\begin{figure}[t]
\includegraphics[width=0.95\columnwidth]{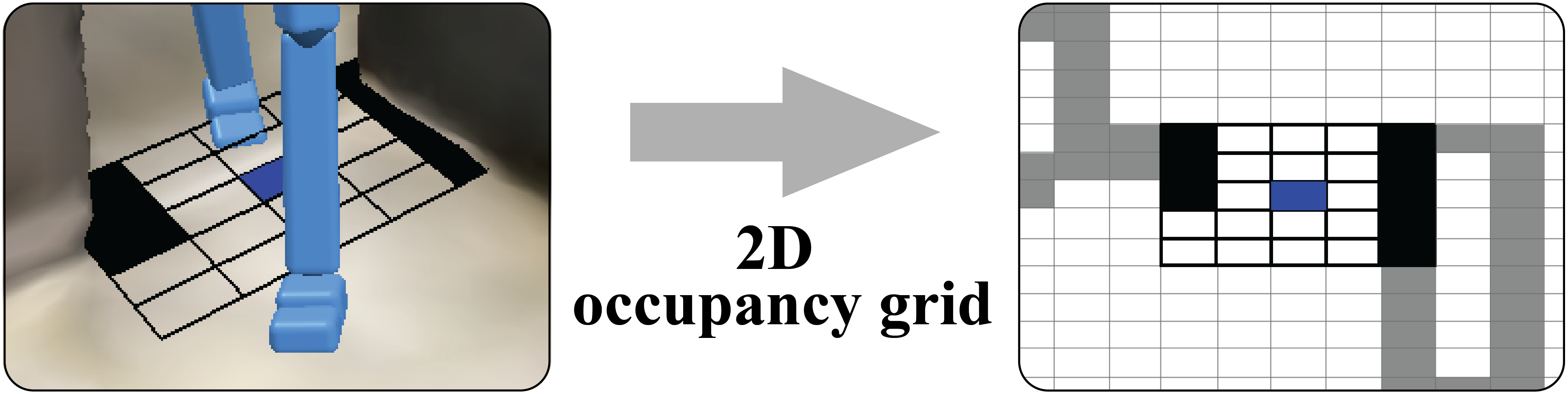}
\caption{Visual representation of the occupancy grid. Grid on the right represents top view. Gray and black are occupied cells while blue indicates the root.}
\label{fig:scenegrid}
\vspace{-.2in}
\end{figure}

\textbf{Motion searching and updating.} 
Given the query motion feature $\mathbf{x}_t$ and the motion features $\mathbf{y}_k$ in the motion database (where $k$ is the index of the clip), motion searching finds the best matches $k^*$ in the motion database by computing the weighted euclidean distances between the query feature and motion database features:
\begin{equation}
k^*=\underset{k}{\arg \min } || \mathbf{w}_f^T (\mathbf{x}_t- \mathbf{y}_k)||^2,
\label{eq:matching_dist}
\end{equation}
where $\mathbf{w}_f$ is a fixed weight vector to control the importance of feature elements.
%
After finding the best match $\hat{\mathbf{m}}_{k^*}$ from the motion database, the motion synthesizer updates it with the predicted motion offset $\mathbf{a}^{offset}_{t}$ from $\mathbf{a}_t$, that is $ \tau(\hat{\mathbf{m}}_{k^* + 1}, \mathbf{a}_{offset})= \mathbf{m}_{t+1}$, where $\hat{\mathbf{m}}_{k^* + 1}$ is the next plausible character posture and $\tau$ is an update function to update selected joints in $\hat{\mathbf{m}}_{k^* + 1}$. In practice, motion searching is performed periodically (e.g., every N-th frame) to make the synthesized motion temporally more coherent.  

 \subsection{Optimizing Scene-Aware Action Controller} \label{sec::learnac}
\begin{figure}[t]
\includegraphics[width=0.95\columnwidth, trim={0 0.4cm 0 0.4cm},clip]{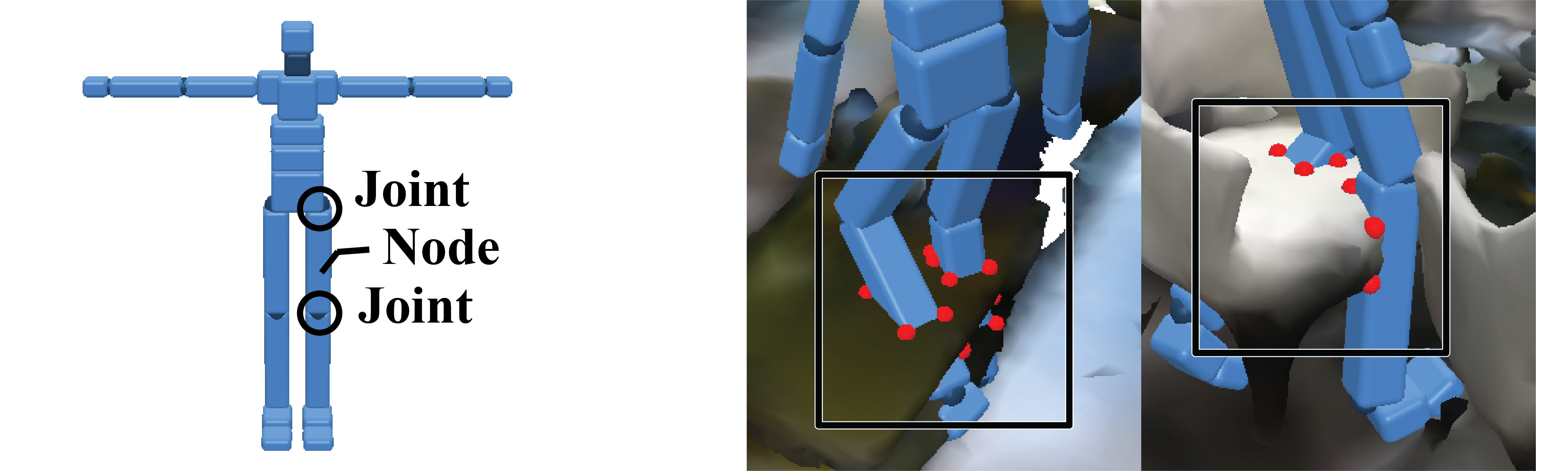}
\caption{(a) Skeleton with joints and box nodes. (b) Automatically detected collision points (colored as red).}
\label{fig:skel}
\vspace{-.2in}
\end{figure}

The objective of our reinforcement learning framework is to optimize the policy by maximizing the discounted cumulative reward. In our method, we design the rewards to guide the character to perform both locomotion and desired actions (e.g., sitting) under common constraints (e.g., smooth transitions, and collision avoidance). 
Our reward function consists of the following terms:
\begin{equation}
 R_\mathrm{total} = w_\mathrm{tr}R_\mathrm{tr} + w_\mathrm{act}R_\mathrm{act} + w_\mathrm{reg}R_\mathrm{reg},
\end{equation}
where $w_\mathrm{tr}$, $w_\mathrm{act}$, and $w_\mathrm{reg}$ are the weights to balance among reward terms. 
The trajectory reward $R_\mathrm{tr}$ is obtained when the character moves towards action $\mathbf{\phi}_A$ while meeting the spatial constraints from the 3D scene, described below:  
\begin{align}
    R_\mathrm{tr} &=  r_\mathrm{coli} \cdot  r_\mathrm{pos} \cdot r_\mathrm{root},  \, \, \text{where}\\
    r_\mathrm{coli} &= \mathrm{exp} \left( -\frac{1}{\sigma_\mathrm{coli}^2}\sum_{b \in \mathbf{B}} w_b \rho(b, \mathbf{W}) \right), \\
  r_\mathrm{pos} &= \mathrm{exp}\left(-\frac{1}{\sigma_\mathrm{root}^2}\sum_{j \in \mathbf{J}}\| \mathbf{p}_{0} -  \mathbf{q}_j\|^2 \right), \\
    r_\mathrm{vel} &= \begin{cases}
1 &\text{when $\dot{\mathbf{p}}_{root} \geq \sigma_{th}$}\\
\sigma_{vel}\|\dot{\mathbf{p}}_{0}\|^2 &\text{else}.
\end{cases}
\end{align}
The collision-avoidance reward $r_\mathrm{coli}$ penalizes collisions with 3D scenes. As depicted in Fig.~\ref{fig:skel} (a), body limbs in the skeletal structure are represented as a set of box-shaped nodes $\mathbf{B}$ with a fixed width, where each element $b \in \mathbf{B}$ is a 3D box representation of legs and arms (we exclude torso and head). The function $\rho(b, \mathbf{W})$ detects the collision between edges of a box-shaped node $b$ with 3D scene $\mathbf{W}$ and returns the number of intersection points. (Fig.~\ref{fig:skel} (b)). $w_b$ is the weight to control the importance of each limb $b$. The collision-avoidance reward is maximized when no penetration occurs, enforcing the policy $\pi$ to generate adequate action $\mathbf{a}_t$ to avoid physically implausible penetrations. $r_\mathrm{pos}$ is obtained when the agent navigates toward targeting action cue $\phi_A$.
$r_\mathrm{vel}$ encourages the character to move by penalizing when the root velocity $\dot{p}_{root}$ is less than a threshold $\sigma_{th}$. $\sigma_\mathrm{coli}$, $\sigma_\mathrm{root}$, and $\sigma_\mathrm{vel}$ are weights to control balance.

Action reward $R_\mathrm{act}$ encourages to fulfill the given action cue $\phi_{A}= \{ \mathbf{q}_{root}, \mathbf{r}_{root}, \mathbf{q}_{rFoot}, \mathbf{q}_{lFoot} \}$: 
\begin{equation}
\begin{split}
    R_\mathrm{act} &=  r_\mathrm{inter} \cdot  r_\mathrm{\Delta t} \cdot r_\mathrm{\Delta v},  \hspace{3mm}\text{where}\\
      r_\mathrm{inter} &= \mathrm{exp}\left(-\frac{1}{\sigma_\mathrm{inter}^2}\sum_{j \in J_A }\| \mathbf{p}_{j} -  \mathbf{q}_{j}\|^2\right), \\
    r_\mathrm{\Delta t} &= \mathrm{exp}\left(-{\sigma_\mathrm{\Delta t}^2}C_\mathrm{tr} \right), 
    r_\mathrm{\Delta v} = \mathrm{exp}\left(-{\sigma_\mathrm{\Delta v}^2}C_\mathrm{vel}\right) \\
\end{split}
\end{equation}
where interaction reward term $r_\mathrm{inter}$ is given when the character switches from navigation to corresponding action to $\phi_A$ and is maximized when the performed action meets the positional constraints of  $\phi_A$. Smoothness reward terms $r_\mathrm{\Delta t}$ and $r_\mathrm{\Delta v}$ minimize the transition cost, which is based on the subpart of the feature distances defined in Eq.~\ref{eq:matching_dist}, where $C_{tr}$ is the weighted feature distances of $p_j$, $\theta_{up}$, and $c$, and $C_{vel}$ is from $\dot{p}$. These are intended to discourage the character from making abrupt changes. 

Regularization reward $R_\mathrm{reg}$ penalizes the $\mathbf{a}^{offset}_t$ excessively modifying the original posture searched in the motion database of $\mathbf{S}$, denoted as $\hat{\mathbf{m}}_t$, and maintains temporal consistency among frames.
\begin{equation}
      R_\mathrm{reg} = \mathrm{exp}\Big(-\frac{1}{\sigma_\mathrm{reg}^2}\Big(\| \hat{\mathbf{m}}_t  - \mathbf{m}_t  \|^2 + \| \mathbf{m}_t  - \mathbf{m}_{t-1} \|^2\Big)\Big). \\ \nonumber
\end{equation}
As reported in~\cite{lee2019scalable, park2019icc}, multiplying rewards with consistent goals can enforce all reward terms to be simultaneously met.
We also use early termination~\cite{peng2018deepmimic} and limited action transitions to accelerate learning. Details are in supp. mat.

\subsection{Generalizing Action Controller} \label{sec:reuse}
While our major focus of the use of RL is for a test-time optimization given a single target task, the optimized policy can handle variations of the task to some extent, as an advantage of the nature of RL. As shown in our experiments in Sec.~\ref{sec:genexp}, we demonstrate that our optimized controller can be directly used for various action cues $\phi_A$ and initials $\mathbf{g}_{init}$ without further optimization for the same scene $\mathbf{W}$.

As an extension of our framework, we can make our controller more generalized by optimizing the policy with random variations of inputs, $\mathbf{g}_{init}$ and $\phi_A$ per each episode during policy optimization. 
This procedure is more similar to the usual RL framework, where the policy is ``learned'' in advance for the target scene $\mathbf{W}$, and applied to the provided inputs during inference. We also demonstrate that our controller can handle a wider range of input variations via this augmentation process. 
%
This extension of our framework can provide better efficiency for the cases where varying tasks are instructed under a fixed 3D scene $\mathbf{W}$. As shown in Sec.~\ref{sec:genexp}, via the generalization process we can directly use the policy for diverse inputs without further optimization. 
Or, if necessary, efficiently fine-tuning the policy is also possible. 
Note that this extension still differs from other learning-based methods~\cite{wang2021synthesizing, hassan2021samp} in that we do not require any scene-paired motion datasets or other supervision.

\subsection{Task-Adaptive Motion Editing}  \label{sec:editing}
\begin{figure}[t]
\includegraphics[width=1.0\columnwidth, trim={0 0cm 0 0.4cm},clip]{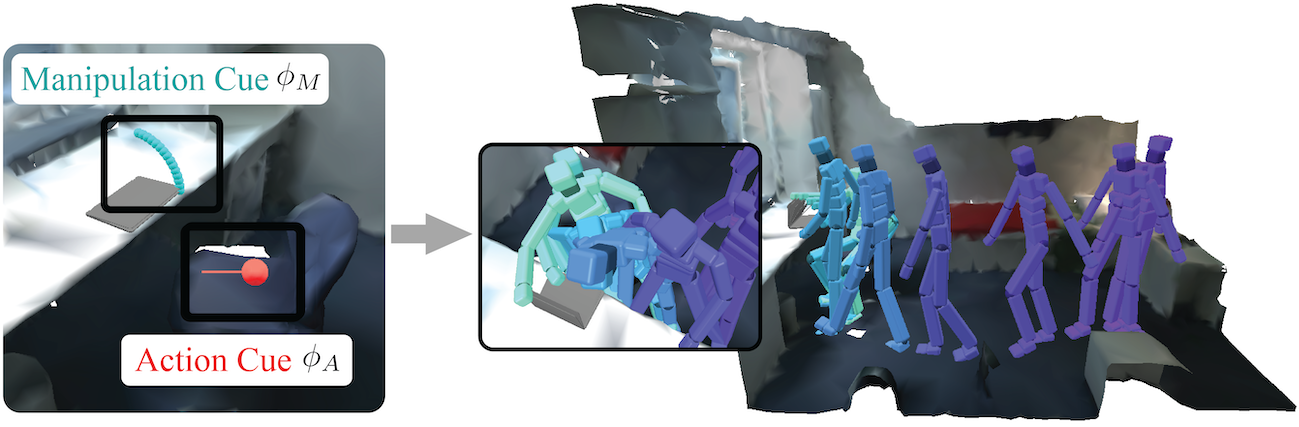}
\caption{Visual representation of system input $\Phi, \mathbf{W}$ and output $\Tilde{\mathbf{M}}$. On the left, action cue $\phi_A$ and manipulation cue $\phi_M$ are shown as red and cyan, respectively. The right is the synthesized motion $\Tilde{\mathbf{M}}$.}
\label{fig:ic}
\vspace{-.2in}
\end{figure}

To cover the diversity in interactions, we include a task-adaptive motion editing module in our motion synthesis framework. 
%
In particular, in the case of object manipulation, manipulation cue $\phi_M$ is provided to enforce an end-effector (e.g., a hand) to follow the desired trajectory expressing the manipulation task on the target object, as in Fig~\ref{fig:ic}. 
The manipulation cue $\phi_M$ can be provided via any possible way, and in our experiments we produce it semi-automatically. 
We compute the desired trajectory by simulating the target articulated object's motion~\cite{xiang2020sapien} by considering a contact point on the surface of the object mesh. 

Not only the edited motion $\Tilde{\mathbf{M}} = E(\mathbf{M})$ should fulfill the sparsely given positional constraints, it should also preserve the temporal consistency and spatial correlations among joints to maintain its naturalness.
We adopt the motion manifold learning approach with convolutional autoencoders~\cite{holden2016deepsynthesis} to compress motion to a latent vector within a motion manifold space. Motion editing is done by searching an optimal latent vector among the manifold. For training the autoencoder, motion sequence, which we denote as $\mathbf{X}$ converted from $\mathbf{M}$, is represented as a time-series of postures by concatenating joint rotations in 6D representations~\cite{zhou20196d}, root height, root transform relative to the previous frame projected on the XZ plane, and foot contact labels. The encoder and decoder module are trained based on reconstruction loss, $||\mathbf{X} - \Psi^{-1}( \Psi \left(\mathbf{X}) \right) ||^2$, where $\Psi$ is the encoder and $\Psi^{-1}$ is the decoder. 

The latent vector from the encoder $\mathbf{z} = \Psi(\mathbf{X})$ from the motion manifold space preserves the spatiotemporal relationship among joints and frames found in natural human motions. As demonstrated in \cite{holden2016deepsynthesis}, editing motions within the manifold space ensures the edited motion to be realistic and coherent. The optimal latent vector $\mathbf{z}^*$ is found by minimizing a loss function $\mathcal{L}$ by constraining the output motions to follow the manipulation constraint $\phi_M$. We also include additional regularizers in $\mathcal{L}$ so that the output motion can maintain the foot locations and root trajectories of the original motion. See supp. mat. for more details on $\mathcal{L}$. Finally, the edited motion $\Tilde{\mathbf{M}}$ can be computed via $\Psi^{-1}(\mathbf{z}^*)$.

\section{Experiments}\label{sec:exp} 
We evaluate LAMA's ability on synthesizing long-term motions in real-world 3D scenes with various human-scene and object interactions involved. We exploit an extensive set of quantitative metrics and perceptual studies for evaluation.

\begin{figure*}[!t]
\centering
    \includegraphics[width=0.98\linewidth,  trim={0 0 0 0},clip]{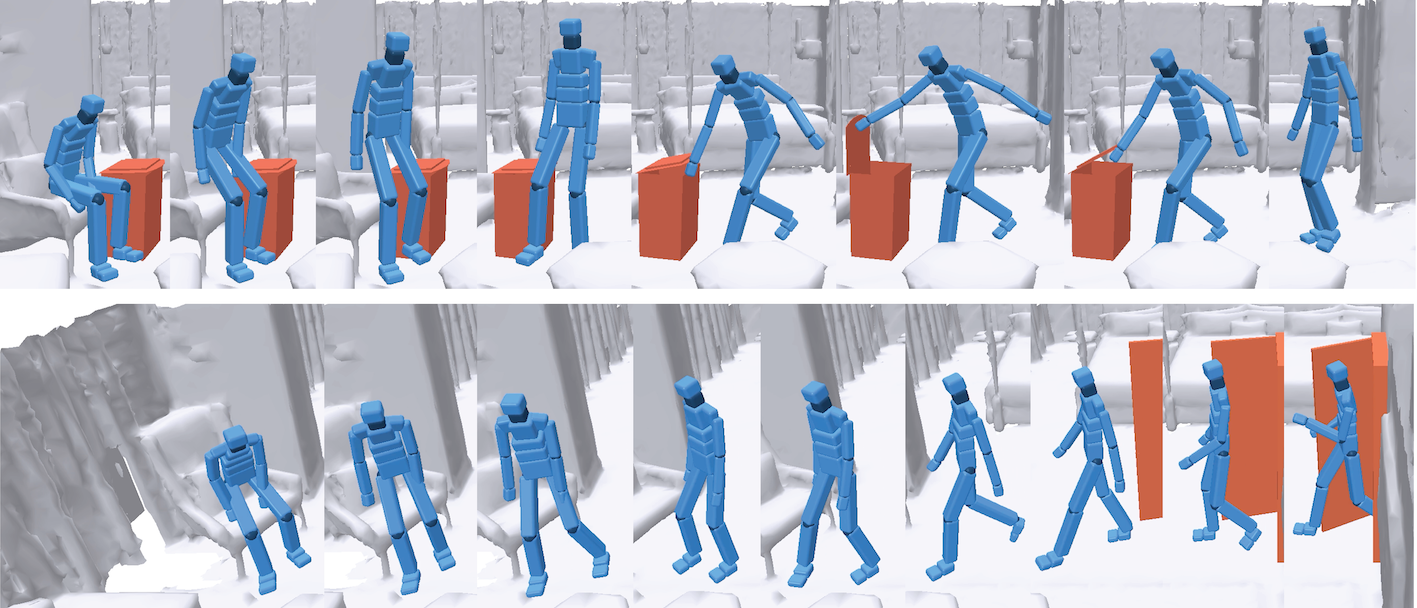}
\caption{Examples of motions which include locomotion, action, and manipulation. Top: opening, closing a trash can lid and sitting on a chair. Bottom: opening door and sitting on a chair.}
\label{fig:result_overall}
\vspace{-.05in}
\end{figure*}

\textbf{Dataset.}  
For constructing the database for the motion synthesizer, we capture a new motion capture dataset involving locomotion and action. Motion is captured with IMU-based system XSens MVN Link~\cite{xsens}. The collected data include high quality human motion with locomotion and interaction in various scenarios, such as walking around at different angles and sitting on a chair with random starting points. 
Captured motion data are post-processed to be suitable for motion matching. 
All the data used in this system are motion capture data (in bvh format) with no scene or object related prior information. We use PROX~\cite{hassan2019prox} and Matterport3D~\cite{matterport3d} datasets for 3D scenes and SAPIEN~\cite{xiang2020sapien} object meshes for manipulation. 
See supp. mat. for details.


\subsection{Experimental Setup}
\noindent\textbf{Evaluation metrics.} As our system does not rely on supervision for motion synthesis, quantifying synthesized quality is challenging due to the lack of ground-truth data or official evaluation metrics. We try to evaluate in terms of physical plausibility and naturalness. 
 
\noindent $\sbullet$  \textbf{Physical Plausibility:} 
 We use contact and penetration metrics to evaluate the physical plausibility of the synthesized motions. Contact penalizes the foot movement when the foot is in contact. 
Since foot contact is a critical element in dynamics, the contact-based metric is closely related to determining the physical plausibility of motions. Penetration loss (``Penetration" in Table~\ref{tab:quan-baseline}) measures implausible cases when the body penetrates the objects in the scene. We compute penetration metric by counting frames where intersection points (Sec.~\ref{sec:ac}) go over a certain threshold.~\footnote{10 for legs and 7 for arms} 
 
\noindent $\sbullet$ \textbf{Naturalness:} We evaluate the naturalness of the synthesized motion via perception study (A/B test) on Amazon Mechanical Turk. The motions used for testing are rendered with the exact same view and 3D characters, making them indistinguishable from the appearance side. Human observers are asked to choose a more natural motion based on two criteria: (1) the character movement is human-like and (2) the movement is plausible in the given scene. Details of the study setup are in supp. mat. 
\begin{table}
    \centering
    \begin{tabular}{lccc}
\toprule
 Method  & Contact & Penetration & Naturalness \\
 \midrule
Wang et al.~\cite{wang2021synthesizing}   & 6.32 & 2.75 & 16.27\\
Wang et al.~\cite{wang2021synthesizing}* & 22.98 & 14.73  & -\\
SAMP~\cite{hassan2021samp}  & 11.75 & 7.18 & 42.04 \\
LAMA (ours) & \textbf{4.34} & \textbf{1.30} & \textbf{100} \\
\bottomrule
\end{tabular}
\caption{\textbf{Baseline Comparison} Foot contact (cm, $\downarrow$) averaged over all frames and penetration (percentage, $\downarrow$) score. Naturalness score (percentage, $\uparrow$) indicates selection ratio relative to LAMA (for LAMA, set to 100).
Wang et al. with an asterisk indicates without post-processing.}
\vspace{-.1in}
\label{tab:quan-baseline}
\end{table}

\medskip \noindent\textbf{Baselines.}
We compare LAMA with the state-of-the-art methods as well as variations of ours. 

\noindent $\sbullet$ \textbf{Wang et al.}~\cite{wang2021synthesizing} is the state-of-the-art long-term motion synthesis method for human-scene interactions within a given 3D scene. 
We use the author's code for evaluation. 
As Wang et al. post-processes synthesized motion to improve foot contact and reduce collisions which are directly related to our metric, we both compare Wang et al. with and without post-processing.
 
\noindent $\sbullet$ \textbf{SAMP}~\cite{hassan2021samp} generates interactions that can be generalized not only for object variations but also random starting points within a given 3D scene. SAMP explicitly exploits path planning to navigate through cluttered 3D scenes. 
 
\noindent $\sbullet$ \textbf{Ablative Baselines} We perform ablation studies on the action controller and motion editing module. 
We perform ablation studies on the scene reward $r_{coli}$, and action offset $\mathbf{a}_t^{offset}$ to present the contribution of both terms on generating scene-aware motions. We also compare our method without the transition reward  $r_{\Delta t}$ and $r_{\Delta v}$ terms (Sec.~\ref{sec:ac}) of the action controller. Finally, we demonstrate the strength of our motion editing module to edit motions naturally (Sec.~\ref{sec:editing}) by comparing it with inverse kinematics (IK).



\subsection{Comparisons with Previous Work}

\begin{figure}[t]
\includegraphics[width=1.0\columnwidth]{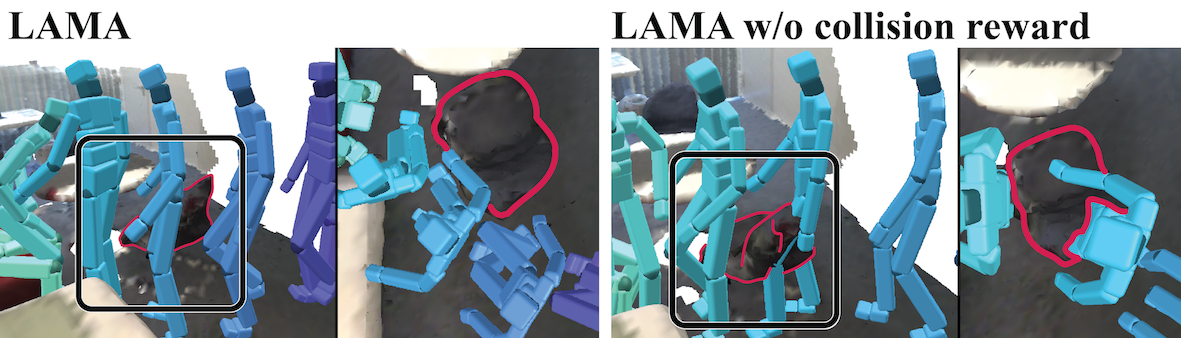}
\caption{Comparison with LAMA (left) and LAMA without collision reward (right). Without collision reward the character fails to avoid collisions with obstacles (red).}
\label{fig:ac-collision}
\vspace{-.1in}
\end{figure}

\begin{figure}[t]
\includegraphics[width=1.0\columnwidth]{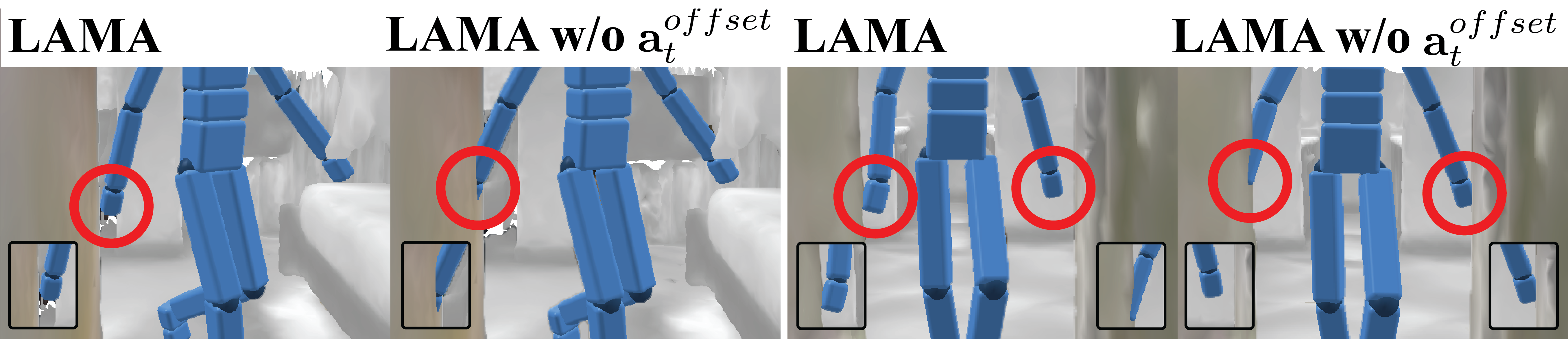}
\caption{Comparison with LAMA (left) and LAMA without action offset (right). The character in original LAMA moves forward while tilting its arms to avoid collision with walls, while in LAMA without action offset does not.}
\label{fig:ac-offset}
\vspace{-.1in}
\end{figure}



\textbf{Evaluation Setup and Details.} For comparison with baselines, we generate 50 motion sequences in total with random input $\mathbf{g}_{init}$ and $\phi_A$ from 4 PROX 3D scenes used in testing for Wang et al.~\cite{wang2021synthesizing}. Since our method is based on test-time optimization without explicit training and testing split, our action controller is optimized per each input, and no prior information on inputs is given before policy optimization. 
It takes 4 to 20 minutes to optimize a policy and 3 to 4 minutes (500 epochs) for optimization in the motion editing module. 
We only consider locomotion and action (walk-to-sit) motions and do not include manipulation as the baselines do not tackle manipulation. 
Contact metric is measured by the position difference of foot in contact, where contact is automatically labeled based on foot velocity.
To compute penetration metric in a fair way, SMPL-X outputs of Wang et al. and SAMP are converted to box-shaped skeletons as in ours and intersection points are counted.
%
Table~\ref{tab:quan-baseline} shows the results.

\textbf{Physical Plausibility.} As shown, LAMA outperforms both Wang et al. and SAMP in physical plausibility. Wang et al. post-processes the synthesized motion to ensure contact and reduce penetration, yet LAMA still outperforms. Moreover, our RL-based method with motion matching shows its advantage in collision avoidance in cluttered 3D scenes compared to path-planning based navigation in SAMP.

\textbf{Naturalness.}
For perception study, we build two separate sets for comparison with Wang et al. and SAMP, 
and each testset is done with non-overlapping participants. For 50 motion sequences per set, 5 unique responses are collected per sequence for comparison. With Wang et al., LAMA received 215 votes while Wang et al received 35 (relative ratio 16.27\%). With SAMP, LAMA received 176 votes, SAMP received 74 (relative ratio 42.04\%).  The results demonstrate that our method greatly outperforms baselines in terms of naturalness as well.



\begin{figure}[t]
    \includegraphics[width=1.0\columnwidth]{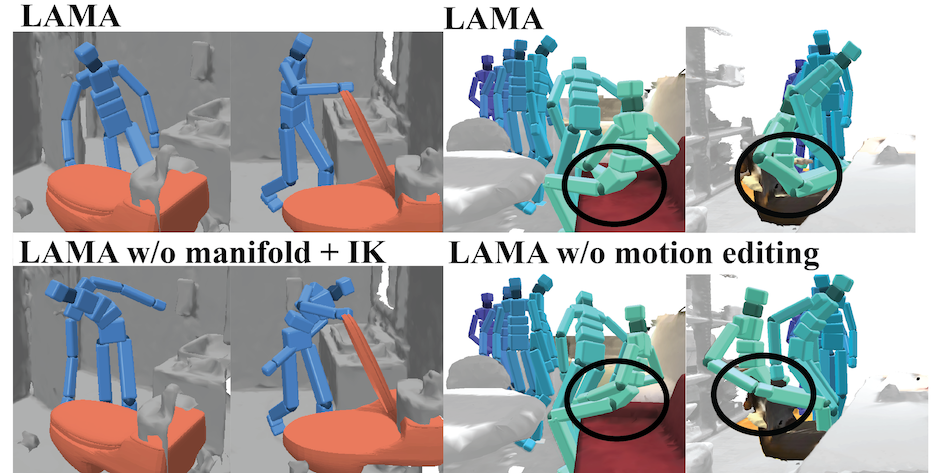}
\caption{(a) Comparison with LAMA (top) and LAMA without manifold and replaced with IK (bottom) of a character opening the toilet lid. (b) Comparison with LAMA (top) and LAMA without motion editing (bottom) in sitting.}
\label{fig:editing}
\end{figure}

\begin{figure}[t]
    \includegraphics[width=1.0\columnwidth,  trim={0.1 0.7cm 0.1 1.0cm},clip]{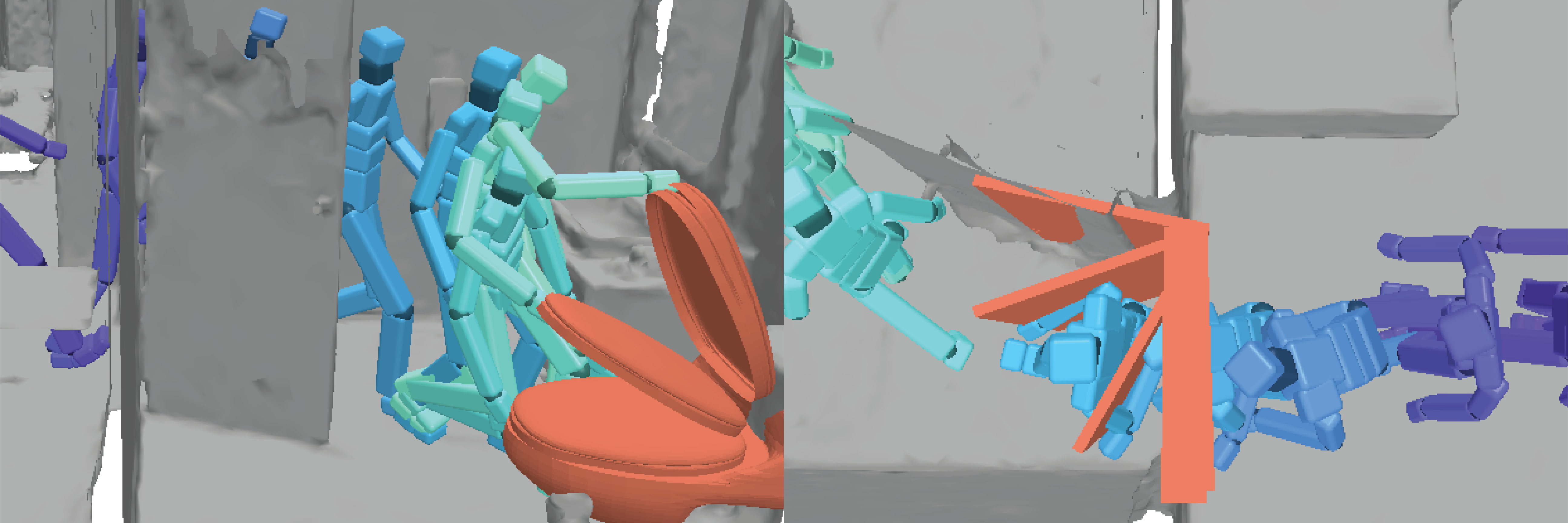}
\caption{Examples of synthesized manipulation motions. The target object for manipulation is colored as orange, the character purple at start and aqua at the end. Left: walking and opening a toilet lid. Right: walking and opening doors.}
\label{fig:result}
\vspace{-.1in}
\end{figure}

\subsection{Ablation Studies }

\textbf{Ablation Studies on Action Controller.}
For quantitative ablations, we compare the original LAMA and the LAMA without collision reward $r_{coli}$. Ablation studies are performed in 5 PROX scenes. In original LAMA, penetration occurs in only \textbf{1.1\%} of the frames among the whole motion sequence, while the ratio is \textbf{15.7\%} in LAMA without $r_{coli}$. The result supports that the $r_{coli}$ enforces the action controller to synthesize motions according to the given 3D scene. Example results are shown in Fig.~\ref{fig:ac-collision}.
We also qualitatively compare the contribution of other components in the action controller. As seen in Fig.~\ref{fig:ac-offset}, without action offset $\mathbf{a}^{offset}_t$ the character does not tilt its limbs to avoid penetration with objects or walls, as the raw motion brought from the motion database does not have any information about the scene. This shows that  $\mathbf{a}^{offset}_t$ also plays a role in generating detailed scene-aware poses. Moreover, the results without smoothness rewards $r_{\Delta t}$ and $r_{\Delta v}$ are not smooth enough, showing unnatural and abrupt movements. 

\textbf{Ablation Studies on Task-Adaptive Motion Editing.} 
We ablate our motion editing module by replacing it with an alternative approach via IK. 
Same as $\phi_M$, only the trajectory of a joint in contact (e.g., the right hand) is given to the IK solver.
As shown in Fig.~\ref{fig:editing} (left), LAMA with motion editing module shows natural moves such as bending knees and tilting hips to make contact. However, results with IK show awkward poses as such spatiotemporal correlations in natural human motions are not reflected in the IK solver. 
Furthermore, 
as seen in Fig.~\ref{fig:editing} (right), the motion editing module makes the character properly sit in chairs with different shapes.

\begin{figure}[t]
    \includegraphics[width=1.0\columnwidth]{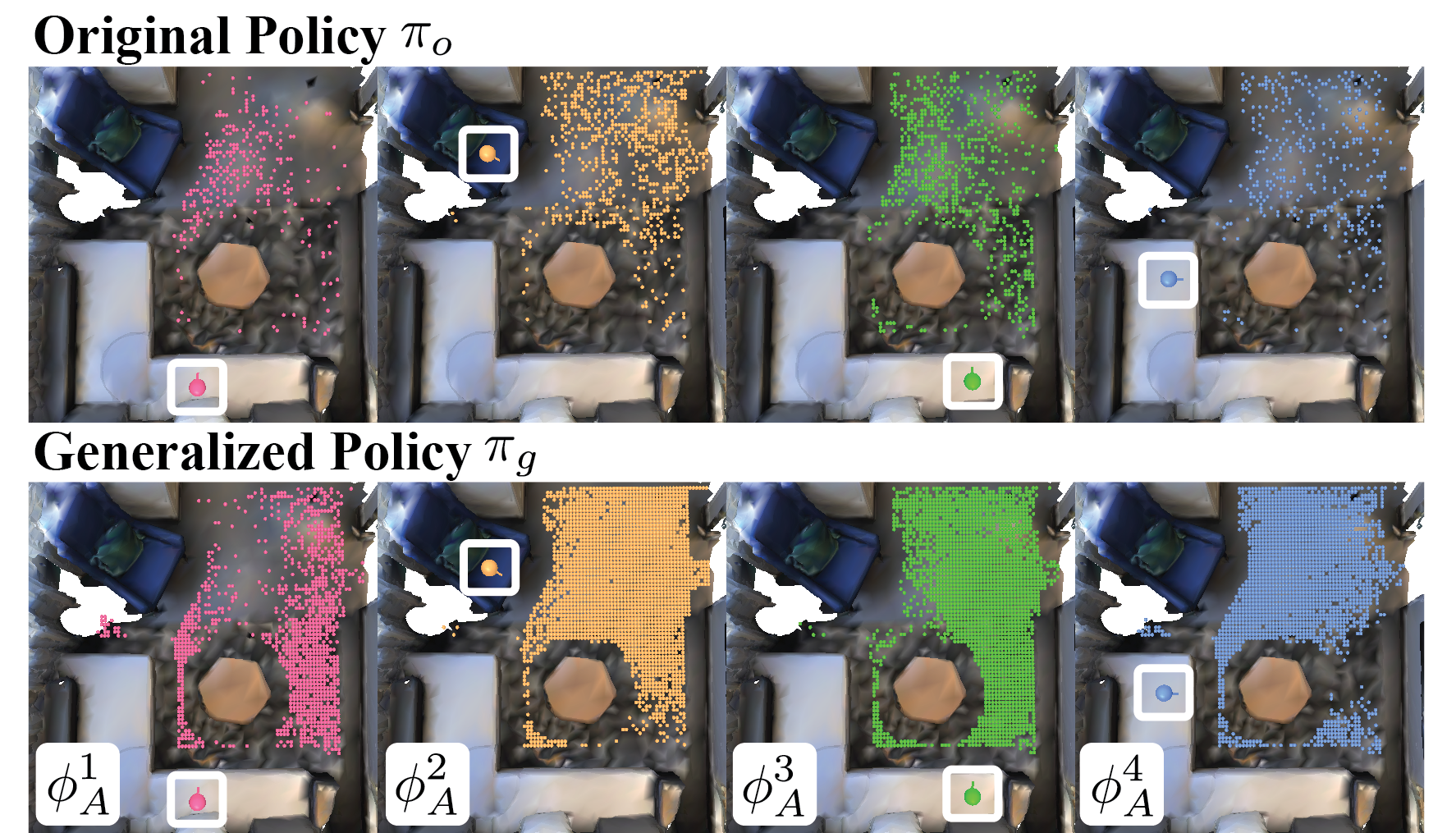}
\caption{Visualization of the range of a policy can cover, with and without generalization. Colored points indicate initial starting point $\mathbf{g}_{init}$ where the policy can synthesize motions meeting the action cue (white).} 
\label{fig:augment}
\vspace{-.1in}
\end{figure}


\subsection{Robustness Test of Action Controller} \label{sec:genexp}
As described in Sec.~\ref{sec:reuse}, utilizing RL for test-time optimization allows the optimized policy to handle variations in input.
In this experiment, we aim to measure the extent to which a policy optimized for a single task $\phi_A$ and initial $\mathbf{g}_{init}$ can generalize to varying inputs. 
To test the robustness with varying initials and tasks, we apply the optimized policy to all possible input variations in the scene and count the number of inputs the policy succeeds in synthesizing. 
From all possible initials sampled, the colored points in Fig~\ref{fig:augment} illustrate initial starting locations where the policy can synthesize motions meeting the given action cue $\phi_A$. 
As shown in Table~\ref{tab:robust-test}, a policy initially optimized for a single set of inputs (red in Fig.~\ref{fig:augment}) can successfully synthesize motions even with distinct set of inputs without any additional optimization. Furthermore, to test the robustness of the generalized policy (described in Sec.~\ref{sec:reuse}), we perform the same test with the policy trained with our augmentation strategy during optimization. As shown above, it shows even more robustness in variations as expected. 

\begin{figure}
    \centering
    \includegraphics[width=1.0\linewidth, trim={0 0cm 0 0cm},clip]{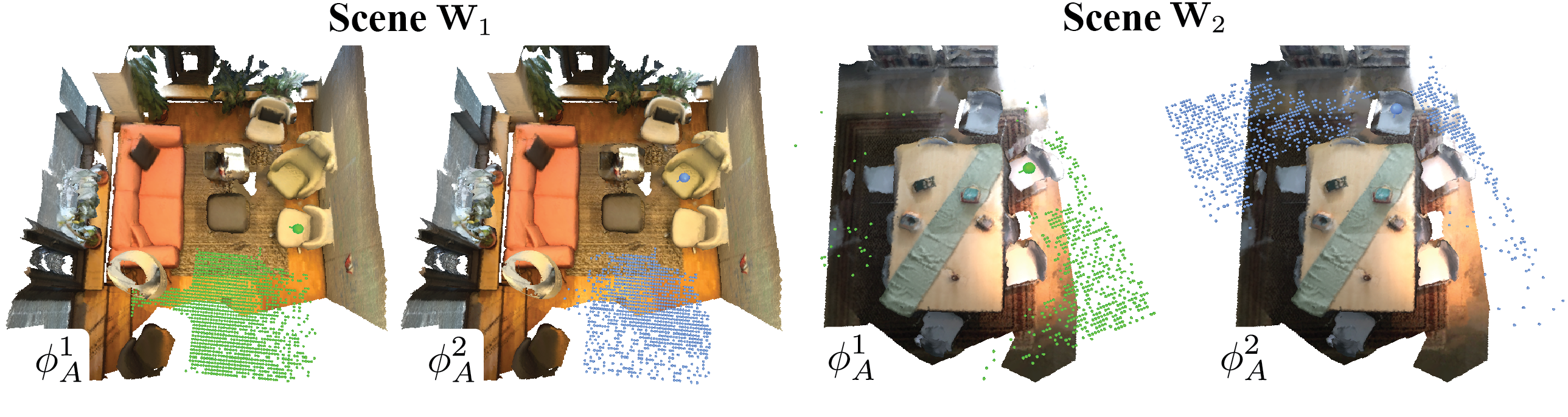}
    \caption{Robustness test on unseen scenes $\mathbf{W}_1$ and $\mathbf{W}_2$. Colored points represent initials where the policy can synthesize motions meeting the action cue.}
    \label{fig:robust-test-scene}
\end{figure}

We further demonstrate the generalization ability among unseen scenes with a policy optimized with the augmentation strategy. The generalized policy (Sec.~\ref{sec:reuse}) optimized in scene $\mathbf{W}_0$ (scene in Fig.~\ref{fig:augment}) are tested on two unseen scenes $\mathbf{W}_1$ and $\mathbf{W}_2$ from PROX~\cite{hassan2019prox} shown in Fig.~\ref{fig:robust-test-scene}. As demonstrated in Table~\ref{tab:robust-test-scene}, an generalized policy (Sec.~\ref{sec:reuse}) optimized with scene $\mathbf{W}_0$ can be generalized to some extent to scene $\mathbf{W}_1$, as $\mathbf{W}_0$ and $\mathbf{W}_1$ shares a similar structure (sofa and chairs around a table). However, as expected, the generalization ability decreases when tested on a totally distinct scene $\mathbf{W}_2$.

Note that the inference time here is about 0.2-3 sec per input as no further policy optimization is required. Details of the test setup are in supp. mat. 

\begin{table}
    \centering
    \begin{tabular}{lcccc}
\toprule
 Method  & $\phi_A^1$ & $\phi_A^2$ & $\phi_A^3$ & $\phi_A^4$ \\
 \midrule
Original Policy & 10.5\% & 25.9\% & 25.5\% & 13.9\% \\
Generalized Policy & 40.2\% & 81.7\% & 77.0\% & 71.7\% \\
\bottomrule
\end{tabular}
\vspace{-.05in}
\caption{\textbf{Robustness Test.} Ratio of $\mathbf{g}_{init}$ (percentage, out of total valid $\mathbf{g}_{init}$ within the scene) which the optimized policy succeeds in synthesizing motion fulfilling $\phi_A^n$. }
\label{tab:robust-test}
\end{table}

\section{Discussion}
We present a unified framework to synthesize human motions within complex real-world 3D scenes with motion-only datasets. We formulate it as a test-time optimization, leveraging RL with motion matching for realistic motion synthesis, and also utilize motion manifold to further cover the diversity of manipulation behaviors.
Our method has been thoroughly evaluated in diverse scenarios, outperforming  previous approaches~\cite{wang2021synthesizing, hassan2021samp}. 

Despite RL is used for test-time optimization, a single policy can cover variations in input and can also be generalized for extensive variations. Combining this framework with supervised learning for further efficiency increase can be an interesting future research direction. Furthermore, although we assume a fixed skeleton throughout the system, interaction motions may change depending on the character's body shapes and sizes. We leave synthesizing motions on varying body shapes as future work.

\begin{table}
    \centering
    \begin{tabular}{lcccc}
\toprule
 Method  & $\mathbf{W}_1, \phi_A^1$ & $\mathbf{W}_1, \phi_A^2$ &  $\mathbf{W}_2, \phi_A^1$ &  $\mathbf{W}_2, \phi_A^2$ \\
 \midrule
- & 70.5\% & 53.4\% & 11.9\% & 21.2\% \\
\bottomrule
\end{tabular}
\caption{\textbf{Robustness Test in Unseen Scenes.} Ratio of $\mathbf{g}_{init}$ (percentage, out of total valid $\mathbf{g}_{init}$ within the scene) which the optimized policy succeeds in synthesizing motion fulfilling $\phi_A^n$ in scene $\mathbf{W}_n$.}
\label{tab:robust-test-scene}
\end{table}

\paragraph{Acknowledgements.}
This work was supported by SNU-Naver Hyperscale AI Center, SNU Creative-Pioneering Researchers Program, NRF grant funded by the Korea government (MSIT) (No. 2022R1A2C2092724), and IITP grant funded by the Korea government (MSIT) (No.2022-0-00156 and No.2021-0-01343). H. Joo is the corresponding author.

\appendix
\section{Supplementary Video}
The supplementary video shows the results of our method, LAMA, on various scenarios. In the video, we show our human motion synthesis results on PROX~\cite{hassan2019prox}, Matterport3D~\cite{matterport3d}, and also our own 3D scene scanned by Polycam App~\cite{policam} with an iPad pro. We use SAPIEN~\cite{xiang2020sapien} object meshes to semi-automatically produce manipulation cues, which is also shown in our videos. As shown, our method successfully produces plausible and natural human motions in many challenging scenarios.

While our original pipeline is designed for test-time optimization, in our video we also qualitatively demonstrate the strength of our framework in generalized scenarios by using a single optimized policy in handling different inputs without further optimization. In the video, we also show a policy optimized via our augmentation strategy (Sec. 3.5.) can handle more extensive input variations.

Our supplementary video contains several ablation studies of our method by showing the importance of collision reward $r_\mathrm{coli}$ in Eq.~(4), transition reward ($r_\mathrm{\Delta t}$, $r_\mathrm{\Delta v}$) in Eq.~(8), posture offset $\mathbf{a}^{offset}_t$ in Action Controller (Sec.~3.2), and our motion editing modules (Sec. 3.5) compared to the traditional Inverse Kinematics (IK). We also show the comparison with previous state-of-the arts~\cite{hassan2021samp, wang2022towards, wang2021synthesizing} and demonstrate that our results produce better quality motions with improved collision avoidance performance in complex 3D scenes.

\section{More Details on Experiments}
In this section, we describe further details on our experiments on the Robustness Test of Action Controller (in Sec. 4.4) and the Perception Study (in Sec 4.2). 

\subsection{Robustness Test of Action Controller (Sec. 4.4)}
In Sec. 4.4, Fig 10, and Table 2 of our main paper, we demonstrate that a single policy optimized for a specific input can handle varying target actions $\phi_A$ and initial $\mathbf{g}_{init}$. We describe more details on the experiment in Sec 4.4.
For the experimental setup, we consider all possible variations for the input to test the generalization ability of the policy trained to a specific input. 
Specifically, a set of ``all'' valid initials $\{\mathbf{g}_{init}\}$ is automatically chosen via grid sampling of the floor plane for the locations $\mathbf{p}_{root}^0$, by excluding points occupied by objects, with a random body orientation for $\mathbf{r}_{root}^0$.  For the action target $\{\phi_A\}$, we manually choose multiple plausible locations (e.g., chairs) for the actions. In the test scene $\mathbf{W}_0$ we use in Fig. 10, there exist 2635 plausible initial positions and we consider 4 target action cues shown in the white boxes in Fig.~10. 

The original policy $\pi_o$ (Fig. 10 top) is optimized to a specific input $\mathbf{g}_{init}$ and action cue $\phi_A^1$, marked as red in the top left of Fig.~10. The colored points in Fig.~10 show the locations where the policy $\pi_o$ achieves the goal successfully without any further optimization for the policy. 
For each input pair $\mathbf{g}_{init}$ and $\phi_A^n$, we perform the motion synthesis with the policy $\pi_o$ 5 times. In each trial, the initial body orientation is chosen randomly to provide more variations.
We determine the policy is successful for the current initial location when no early termination conditions (collision, stall, moving out of the scene) are met while fulfilling $\phi_A^n$ at least twice out of the 5 trials. As shown in Fig. 10 and Table 2, our action controller optimized for a specific target can be applicable to many input variations.

We perform the same test for the generalized policy (described in Sec. 3.5) in the bottom of Fig. 10 and Tab. 2. As shown, this policy can cover much more extensive input variations on the same scene.

\begin{table}[t]
\centering
\begin{tabular}{lll}
\toprule
 Name  & Foot Contact & Penetration \\
 \midrule
Single Optimized Policy  & 4.16  & 1.35 \\
Generalized Policy  & 4.47  & 1.41 \\
 \bottomrule
\end{tabular}
\caption{Physical plausibility measurement of the synthesized motion from the robustness test. (Sec 4.4) }
\label{tab:motion_qual_robust}
\end{table}

\textbf{Comparison of Computation Time.}
 As a test-time optimization without requiring scene-paired motion datasets, our original framework takes time to train a policy from a scratch for a given input pair. However, reusing the same policy that is optimized for the specific input for other inputs can greatly reduce the computation time, because no further optimization is needed for the policy.  
To compare the time between performing the inference only and optimizing a policy from scratch, we test with 5 input pairs consisting of initial $\mathbf{g}_{init}$ and $\phi_A$. Here, the term ``inference only" indicates that we use a pre-optimized policy without any further optimization for varying inputs. 
As the result, the inference-only scenario takes \textbf{0.15} seconds on average per input pair for motion synthesis, while optimizing a policy from scratch per pair takes \textbf{6.32} minutes (379 seconds) on average. As shown, the capability of the reinforcement learning framework provides the potential to greatly improve the efficiency of our method. 


\textbf{Motion Quality Measurement.}
We also evaluate the physical plausibility of the synthesized motion in the robustness test in Sec. 4.4. An optimized policy synthesized 15 motion sequences with distinct input pairs (the input pair which the policy is initially optimized to is not included). We also perform the measurement to motions synthesized by the generalized policy optimized with an augmentation strategy. The results are shown in Table~\ref{tab:motion_qual_robust}. This shows while a policy can handle variations in input, there is no performance drop in the synthesized motion quality.






\subsection{Perception Study Setup}
The videos used for perception study are in the supplementary video. We include 3 videos per set to the supplementary video.


\section{More Details on Implementations}
\subsection{Action Controller}
\paragraph{Implementation Details.}
The policy and the value network of the action controller module consists of 4 and 2 fully connected layers of 256 nodes, respectively. The control policy is optimized through Proximal Policy Optimization (PPO) algorithm~\cite{schulman2017proximal}. Adam optimizer~\cite{kingma2014adam} is used with Nvidia RTX 3090 GPU. 
For the action controller $\mathbf{A}$ and motion synthesizer module $\mathbf{S}$, we use  the animation library DART~\cite{lee2018dart}. We also use a publicly available PPO implementation~\cite{park2019icc, lee2021parametric}, where we remove the variable time-stepping functions stepping in \cite{lee2021parametric} by following the original PPO algorithm. 
The details of the optimization regarding the policy and value network of the action controller are written in Table~\ref{tab:controller-parameter}. 

\paragraph{Acceleration Techniques.}
As written in the main paper, we use early termination conditions to accelerate policy optimization. The episode is terminated when (1) the character moves out of the scene bounding box; (2) when the collision reward $r_{coli}$ is under a certain threshold; and (3) the root velocity for a specific time duration (50 frames) is under a certain threshold to prevent the character standing still for a overly long time. Also, the action controller first checks in advance whether the action signal is valid when it makes transitions from locomotion to other actions. When the nearest feature distance of Eq. 2 in the motion synthesizer (Sec. 3.3) is over a certain threshold, the action controller discards the transition and continues navigating. 

\begin{table}[t]
\centering
\begin{tabular}{ll}
\toprule
 Name  & Value \\
 \midrule
 Learning rate of policy network  & 2e-4  \\
 Learning rate of value network  & 0.001  \\
 Discount factor ($\gamma$)  & 0.95 \\
 GAE and TD ($\lambda$)  & 0.95 \\
 Clip parameter ($\epsilon$) & 0.2 \\
 \# of tuples per policy update & 30000 \\
Batch size for policy/value update & 512 \\
 \bottomrule
\end{tabular}
\caption{Details on the hyper-parameters for learning the control policy of the Action Controller $\mathbf{A}$.}
\label{tab:controller-parameter}
\end{table}

\subsection{Motion Synthesizer}
\paragraph{Motion Database Information.}
Motion is captured by an IMU based system XSens MVN Link~\cite{xsens} and is post-processed via XSens MotionCloud software~\cite{xsensmotioncloud}. The captured motion is then retargeted to a single unified skeleton using Autodesk MotionBuilder and is post-processed to be suitable for motion matching. For action motions we mirror the motion segments for data augmentation.
The length (in frames) of motion segments (``Seg. Length'' in tables), number of motion segment (``Seg. Count'' in tables), and the number of total frames (``Total Frames'' in tables) are summarized in Table~\ref{tab:motion-database}.  

\paragraph{Action-Specific Feature Definition.}
The motion feature, as defined in our main paper Sec 3.3, represents both the current state of the motion and a short term future movements: $ f( \mathbf{m}) = \{ \{ p_\mathrm{j} \},  \{ \dot{p}_\mathrm{j} \}, \theta_{up}, c, \mathbf{o}_\mathrm{future} \}$. In particular the action specific feature $\mathbf{o}_\mathrm{future} = \{ \{p_0^{\Delta t} \}, \{r_0^{\Delta t} \} \}$ contains future motions so that the motion search process can take into account the future motion consistency, where $p_0^{\Delta t}, r_0^{\Delta t} \in \mathbb{R}^2$ are the position and orientation of root joint at $\Delta t$ frames later from the current target frame. For locomotion, we extract $\Delta t=10$, 20, and 30 frames in the future (at 30Hz) following \cite{clavet2016motion}, as addressed in our main paper. For sitting, we specifically choose $\Delta t$ as the frame where the character completes the sit-down motion. The major motivation of this design choice is encourage the motion synthesizer to search the motion clips with the desired target action.

\subsection{Motion Editing via Motion Manifold}

\paragraph{Implementation Details for Models and Training.}
The encoder and decoder of the task-adaptive motion editing module consist of three convolutional layers.
For the convolutional autoencoder of task-adaptive motion editing, we use PyTorch~\cite{pytorch}, FairMotion~\cite{gopinath2020fairmotion}, and PyTorch3d~\cite{ravi2020pytorch3d}. The autoencoder is trained with the Adam optimizer~\cite{kingma2014adam} with learning rate 0.0001. We use Nvidia RTX 3090 GPU.
We use 3 layers of 1D temporal-convolutions with kernel width of 25 and stride 2, and the channel dimension of each output feature is 256.
For training the autoencoder module in task-adaptive motion editing we use data in Mixamo~\cite{mixamo}, Lafan1~\cite{harvey2020robust}, COUCH~\cite{zhang2022couch}, and ours. The training datasets are summarized in Table~\ref{tab:manifold-data}. Note that data used for training the autoencoder also does not include scene related information (in bvh format), and we use different pre-processing steps between the Motion Editing module and the Motion Synthesizer.



\paragraph{Reconstruction Loss.} The encoder $\Psi$ and decoder $\Psi^{-1}$ are trained based on reconstruction loss $\mathcal{L}_\mathrm{recon} = ||\mathbf{X} - \Psi^{-1}( \Psi \left(\mathbf{X}) \right) ||^2$, where: 
\begin{equation}
  \begin{split}
\mathcal{L}_\mathrm{recon} &= w_{c}\mathcal{L}_{\textrm{contact}} + w_{r}\mathcal{L}_{\textrm{root}} + w_{q}\mathcal{L}_{\textrm{quat}} + w_{p}\mathcal{L}_{\textrm{pos}}.
  \end{split}    
\end{equation}
$\mathcal{L}_\mathrm{contact}$, $\mathcal{L}_\mathrm{root}$, and $\mathcal{L}_\mathrm{quat}$ are the MSE losses of foot contact labels,  root status (height and transform relative to the previous frame projected on the XZ plane), and the joint rotations in 6D representations~\cite{zhou20196d}.
To penalize errors accumulating along the kinematic chain, we perform forward kinematics (FK) and measure the global position distance of joints between the original and reconstructed motion. As global positions of the joints are highly dependent on the root positions, for the early epochs, the distance is measured based on root-centric coordinates to ignore the global location of roots, which we found empirically more stable.  Also, during training we used an augmentation technique of adding noise to normalized input. Noise is sampled from the normal distribution $\mathcal{N}(0,\,1)$ multiplied with a scale of 0.01. We found the technique empirically increases reconstructed motion quality.


\paragraph{Motion Editing Loss.}
For motion editing, the positional loss and regularization loss are defined as follows.
\begin{equation}
  \begin{split}
\mathcal{L} &= w_{p}\mathcal{L}_{\textrm{pos}} + w_{f}\mathcal{L}_{\textrm{foot}} + w_{r}\mathcal{L}_{\textrm{root}}, \hspace{3mm}\text{where} \\
 \mathcal{L}_\mathrm{pos} &= \sum_{\mathbf{q}^t_j \in \mathrm{\phi}_M}\| \textbf{p}^t_{j} -  \mathbf{q}^t_j\|^2, \\
\mathcal{L}_\mathrm{foot} &= \sum_{foot}\| \mathbf{p}_{foot}^{e} - \mathbf{p}_{foot}^{i}\|^2, \\
\mathcal{L}_\mathrm{root} &= w_r \| \mathbf{r}^{e}_\textrm{xz} -  \mathbf{r}^{i}_\textrm{xz}\|^2 + w_{\Delta r}\| \dot{\mathbf{r}}^{e}_\textrm{xz} -  \dot{\mathbf{r}}^{i}_\textrm{xz}\|^2. \\
  \end{split}    
\end{equation}
$\mathbf{p}_j$ denotes positions of joint $j$, and  $\mathbf{r}$, $\dot{\mathbf{r}}$ denotes root positions and velocities respectively. Superscript $e$ and $i$ indicates whether it is from edited or initial motion, respectively. Subscript $\textrm{xz}$ indicates the vector is projected onto the XZ plane. The loss term $\mathcal{L}$ enforces the edited motion to maintain contact and root trajectory (in the XZ plane) of the initial motion, while generating natural movements of the other joints to meet the sparse positional constraints. For minimizing losses, Adam optimizer~\cite{kingma2014adam} is used as well with a learning rate of 0.005. 


\paragraph{Generating Manipulation Cues from SAPIEN~\cite{xiang2020sapien}.}
While the manipulation cue $\phi_M = [\mathbf{v}(R_t, T_t, \theta_t) ]_t$ can be provided via diverse ways depending on the applications, we mainly consider the scenarios of interacting with articulated objects. For this purpose, we semi-automatically produce the manipulation cues by extracting the desired target vertex trajectories of the parts of articulated objects from the SAPIEN dataset~\cite{xiang2020sapien}. Specifically, we place a target object in our 3D scene, and choose a target vertex $\mathbf{v}$ of the object where we assume the character's hand contacts to manipulate the target part (e.g., a vertex in the lid of a trash can object). Then, the trajectory of the vertex $\phi_M = [\mathbf{v}(R_t, T_t, \theta_t) ]_t$ can be obtained by varying the parameter for the articulated motion $\theta$ with a fixed interval, where $R_t$, $T_t$, are the global orientation and translation of the object and $\theta_t$ is the parameters for the object articulation (e.g., the hinge angle of the cover of a laptop) at time $t$. $\mathbf{v}(\cdot)$ represents the 3D location of the chosen vertex $\mathbf{v}$ given the parameters. The resulting manipulation cue $\phi_M$ is the target trajectory that a hand joint should follow for the manipulation motion.
Note that our system requires only the manipulation cue $\phi_M$, and the 3D object mesh is shown only for visualization purposes, where we visualize it with the synced $\theta_t$.

\paragraph{Further Implementation Details for Manipulation} 
Given the initial motion output $\mathbf{M}$ synthesized from the Action Controller, $\mathbf{M}=\{\mathbf{m}_t\}_{t=1}^T$ and the manipulation cue $\phi_M=\{\mathbf{m}_{t'}\}_{{t'}=1}^{\tau}$, our system is also given the corresponding time segment $[t_i, t_f]$ where we want to edit the motion to follow the manipulation cue (we assume the same duration, i.e., $t_f - t_i = \tau$). Then motion editing is performed on the target motion segment, $\Tilde{\mathbf{M}}_{t_i:t_f} = \mathbf{E}(\mathbf{M}_{t_i:t_f})$, which subsequently replaces the corresponding part in $\mathbf{M}$ to form $\Tilde{\mathbf{M}}$ as the final output. 

Depending on possible applications (e.g, sitting down and opening a laptop), the manipulation motion may need to be ``added'' in the middle or the end of the synthesized motion $\mathbf{M}$. In this case, we simply duplicate the target frame by $\tau$ to build a longer motion $\mathbf{M}=\{\mathbf{m}_t\}_{t=1}^{T+\tau}$, and apply the motion editing to the target motion segment that is a stationary motion produced via the duplication.

\begin{table}
\centering
\begin{tabular}{llll}
\toprule
 Label  & Seg. Length & Seg. Count & Total Frames \\
 \midrule
 Locomotion  & 10 &  23832 & 24267 \\
 Sit  & 50 -- 85  & 6230 & 15130 \\
 \bottomrule
\end{tabular}
\caption{Details on pre-processed motion datasets per each action category of the motion database in $\mathbf{S}$. }
\label{tab:motion-database}
\end{table}


\begin{table}
\centering
\begin{tabular}{ll}
\toprule
 Name  & Value \\
 \midrule
 Motion sequence length & 120 \\
 Number of sequence (training) & 45713 \\
 Number of sequence (test) & 11040\\
 Number of sequence (validation) & 5268 \\
 \bottomrule
\end{tabular}
\caption{Details on pre-processed motion datasets for training our motion editing module $\mathbf{M}$.}
\label{tab:manifold-data}
\end{table}



{\small
\bibliographystyle{ieee_fullname}
\bibliography{egbib}

\begin{thebibliography}{10}\itemsep=-1pt

\bibitem{policam}
Polycam - lidar and 3d scanner for iphone i\& android.
\newblock \url{https://poly.cam/}.

\bibitem{xsensmotioncloud}
Xsens motioncloud.
\newblock
  \url{https://www.movella.com/products/motion-capture/xsens-motioncloud}.

\bibitem{xsens}
Xsens mvn link.
\newblock \url{https://www.movella.com/products/motion-capture/xsens-mvn-link}.

\bibitem{aberman2020skeleton}
Kfir Aberman, Peizhuo Li, Dani Lischinski, Olga Sorkine-Hornung, Daniel
  Cohen-Or, and Baoquan Chen.
\newblock Skeleton-aware networks for deep motion retargeting.
\newblock {\em ACM Trans. Graph}, 39(4), 2020.

\bibitem{aberman2020unpaired}
Kfir Aberman, Yijia Weng, Dani Lischinski, Daniel Cohen-Or, and Baoquan Chen.
\newblock Unpaired motion style transfer from video to animation.
\newblock {\em ACM Trans. Graph.}, 39(4), 2020.

\bibitem{bergamin2019drecon}
Kevin Bergamin, Simon Clavet, Daniel Holden, and James~Richard Forbes.
\newblock Drecon: data-driven responsive control of physics-based characters.
\newblock {\em ACM Trans. Graph.}, 38(6), 2019.

\bibitem{buttner2015motion}
Michael Büttner and Simon Clavet.
\newblock Motion matching - the road to next gen animation.
\newblock In {\em Proc. of Nucl.ai}, 2015.

\bibitem{cao2020long}
Zhe Cao, Hang Gao, Karttikeya Mangalam, Qi-Zhi Cai, Minh Vo, and Jitendra
  Malik.
\newblock Long-term human motion prediction with scene context.
\newblock In {\em ECCV}, 2020.

\bibitem{matterport3d}
Angel Chang, Angela Dai, Thomas Funkhouser, Maciej Halber, Matthias Niessner,
  Manolis Savva, Shuran Song, Andy Zeng, and Yinda Zhang.
\newblock Matterport3d: Learning from rgb-d data in indoor environments.
\newblock In {\em 3DV}, 2017.

\bibitem{chao2021chair}
Yu-Wei Chao, Jimei Yang, Weifeng Chen, and Jia Deng.
\newblock Learning to sit: Synthesizing human-chair interactions via
  hierarchical control.
\newblock In {\em AAAI}, 2021.

\bibitem{clavet2016motion}
Simon Clavet.
\newblock Motion matching and the road to next-gen animation.
\newblock In {\em Proc. of GDC}, 2016.

\bibitem{eom2019mpc}
Haegwang Eom, Daseong Han, Joseph~S Shin, and Junyong Noh.
\newblock Model predictive control with a visuomotor system for physics-based
  character animation.
\newblock {\em ACM Trans. Graph.}, 39(1), 2019.

\bibitem{fragkiadaki2015recurrent}
Katerina Fragkiadaki, Sergey Levine, Panna Felsen, and Jitendra Malik.
\newblock Recurrent network models for human dynamics.
\newblock In {\em ICCV}, 2015.

\bibitem{gopinath2020fairmotion}
Deepak Gopinath and Jungdam Won.
\newblock fairmotion - tools to load, process and visualize motion capture
  data.
\newblock Github, 2020.

\bibitem{habibie2017recurrent}
Ikhsanul Habibie, Daniel Holden, Jonathan Schwarz, Joe Yearsley, and Taku
  Komura.
\newblock A recurrent variational autoencoder for human motion synthesis.
\newblock In {\em BMVC}, 2017.

\bibitem{harvey2020robust}
F{\'e}lix~G Harvey, Mike Yurick, Derek Nowrouzezahrai, and Christopher Pal.
\newblock Robust motion in-betweening.
\newblock {\em ACM Trans. Graph}, 39(4), 2020.

\bibitem{hassan2021samp}
Mohamed Hassan, Duygu Ceylan, Ruben Villegas, Jun Saito, Jimei Yang, Yi Zhou,
  and Michael Black.
\newblock Stochastic scene-aware motion prediction.
\newblock In {\em ICCV}, 2021.

\bibitem{hassan2019prox}
Mohamed Hassan, Vasileios Choutas, Dimitrios Tzionas, and Michael~J. Black.
\newblock Resolving {3D} human pose ambiguities with {3D} scene constraints.
\newblock In {\em ICCV}, 2019.

\bibitem{hassan2021posa}
Mohamed Hassan, Partha Ghosh, Joachim Tesch, Dimitrios Tzionas, and Michael~J
  Black.
\newblock Populating 3d scenes by learning human-scene interaction.
\newblock In {\em CVPR}, 2021.

\bibitem{holden2020learned}
Daniel Holden, Oussama Kanoun, Maksym Perepichka, and Tiberiu Popa.
\newblock Learned motion matching.
\newblock {\em ACM Trans. Graph.}, 39(4), 2020.

\bibitem{holden2017phase}
Daniel Holden, Taku Komura, and Jun Saito.
\newblock Phase-functioned neural networks for character control.
\newblock {\em ACM Trans. Graph.}, 36(4), 2017.

\bibitem{holden2016deepsynthesis}
Daniel Holden, Jun Saito, and Taku Komura.
\newblock A deep learning framework for character motion synthesis and editing.
\newblock {\em ACM Trans. Graph.}, 35(4), 2016.

\bibitem{huang2022behave}
Chun-Hao~P Huang, Hongwei Yi, Markus H{\"o}schle, Matvey Safroshkin, Tsvetelina
  Alexiadis, Senya Polikovsky, Daniel Scharstein, and Michael~J Black.
\newblock Capturing and inferring dense full-body human-scene contact.
\newblock In {\em CVPR}, 2022.

\bibitem{hyun2016motion}
Kyunglyul Hyun, Kyungho Lee, and Jehee Lee.
\newblock Motion grammars for character animation.
\newblock In {\em Computer Graphics Forum}, volume~35, 2016.

\bibitem{jiang2022chairs}
Nan Jiang, Tengyu Liu, Zhexuan Cao, Jieming Cui, Yixin Chen, He Wang, Yixin
  Zhu, and Siyuan Huang.
\newblock Chairs: Towards full-body articulated human-object interaction.
\newblock {\em arXiv preprint arXiv:2212.10621}, 2022.

\bibitem{jiang2022neuralhofusion}
Yuheng Jiang, Suyi Jiang, Guoxing Sun, Zhuo Su, Kaiwen Guo, Minye Wu, Jingyi
  Yu, and Lan Xu.
\newblock Neuralhofusion: Neural volumetric rendering under human-object
  interactions.
\newblock In {\em CVPR}, 2022.

\bibitem{kim2014shape2pose}
Vladimir~G Kim, Siddhartha Chaudhuri, Leonidas Guibas, and Thomas Funkhouser.
\newblock Shape2pose: Human-centric shape analysis.
\newblock {\em ACM Trans. Graph.}, 33(4), 2014.

\bibitem{kingma2014adam}
Diederik~P Kingma and Jimmy Ba.
\newblock Adam: A method for stochastic optimization.
\newblock {\em arXiv preprint arXiv:1412.6980}, 2014.

\bibitem{lee2002interactive}
Jehee Lee, Jinxiang Chai, Paul~SA Reitsma, Jessica~K Hodgins, and Nancy~S
  Pollard.
\newblock Interactive control of avatars animated with human motion data.
\newblock In {\em Proceedings of the 29th annual conference on Computer
  graphics and interactive techniques}, 2002.

\bibitem{lee2018dart}
Jeongseok Lee, Michael~X Grey, Sehoon Ha, Tobias Kunz, Sumit Jain, Yuting Ye,
  Siddhartha~S Srinivasa, Mike Stilman, and C~Karen Liu.
\newblock Dart: Dynamic animation and robotics toolkit.
\newblock {\em The Journal of Open Source Software}, 3(22), 2018.

\bibitem{lee2004precomputing}
Jehee Lee and Kang~Hoon Lee.
\newblock Precomputing avatar behavior from human motion data.
\newblock In {\em Proceedings of the 2004 ACM SIGGRAPH/Eurographics symposium
  on Computer animation}, 2004.

\bibitem{lee2018interactive}
Kyungho Lee, Seyoung Lee, and Jehee Lee.
\newblock Interactive character animation by learning multi-objective control.
\newblock {\em ACM Trans. Graph.}, 37(6), 2018.

\bibitem{lee2021timecritical}
Kyungho Lee, Sehee Min, Sunmin Lee, and Jehee Lee.
\newblock Learning time-critical responses for interactive character control.
\newblock {\em ACM Trans. Graph.}, 40(4), 2021.

\bibitem{lee2006motion}
Kang~Hoon Lee, Myung~Geol Choi, and Jehee Lee.
\newblock Motion patches: building blocks for virtual environments annotated
  with motion data.
\newblock In {\em ACM SIGGRAPH 2006 Papers}. 2006.

\bibitem{lee2022comcon}
Seunghwan Lee, Phil~Sik Chang, and Jehee Lee.
\newblock Deep compliant control.
\newblock In {\em ACM SIGGRAPH 2022 Conference Proceedings}, 2022.

\bibitem{lee2021parametric}
Seyoung Lee, Sunmin Lee, Yongwoo Lee, and Jehee Lee.
\newblock Learning a family of motor skills from a single motion clip.
\newblock {\em ACM Trans. Graph.}, 40(4), 2021.

\bibitem{lee2019scalable}
Seunghwan Lee, Moonseok Park, Kyoungmin Lee, and Jehee Lee.
\newblock Scalable muscle-actuated human simulation and control.
\newblock {\em ACM Trans. Graph.}, 38(4), 2019.

\bibitem{levine2012continuous}
Sergey Levine, Jack~M Wang, Alexis Haraux, Zoran Popovi{\'c}, and Vladlen
  Koltun.
\newblock Continuous character control with low-dimensional embeddings.
\newblock {\em ACM Trans. Graph}, 31(4), 2012.

\bibitem{li2021aict}
Ruilong Li, Shan Yang, David~A. Ross, and Angjoo Kanazawa.
\newblock Ai choreographer: Music conditioned 3d dance generation with aist++.
\newblock In {\em ICCV}, 2021.

\bibitem{ling2020motionvae}
Hung~Yu Ling, Fabio Zinno, George Cheng, and Michiel Van De~Panne.
\newblock Character controllers using motion vaes.
\newblock {\em ACM Trans. Graph.}, 39(4), 2020.

\bibitem{lucas2002motion}
Kovar Lucas, Gleicher Michael, and Pighin Fr{\'e}d{\'e}ric.
\newblock Motion graphs.
\newblock In {\em Proceedings of the 29th Annual Conference on Computer
  Graphics and Interactive Techniques}, 2002.

\bibitem{martinez2017human}
Julieta Martinez, Michael~J Black, and Javier Romero.
\newblock On human motion prediction using recurrent neural networks.
\newblock In {\em CVPR}, 2017.

\bibitem{merel2020catch}
Josh Merel, Saran Tunyasuvunakool, Arun Ahuja, Yuval Tassa, Leonard
  Hasenclever, Vu Pham, Tom Erez, Greg Wayne, and Nicolas Heess.
\newblock Catch \& carry: reusable neural controllers for vision-guided
  whole-body tasks.
\newblock {\em ACM Trans. Graph.}, 39(4), 2020.

\bibitem{mixamo}
Adobe's Mixamo.
\newblock \url{https://www.mixamo.com}, 2017.

\bibitem{park2019icc}
Soohwan Park, Hoseok Ryu, Seyoung Lee, Sunmin Lee, and Jehee Lee.
\newblock Learning predict-and-simulate policies from unorganized human motion
  data.
\newblock {\em ACM Trans. Graph.}, 38(6), 2019.

\bibitem{pytorch}
Adam Paszke, Sam Gross, Francisco Massa, Adam Lerer, James Bradbury, Gregory
  Chanan, Trevor Killeen, Zeming Lin, Natalia Gimelshein, Luca Antiga, Alban
  Desmaison, Andreas Kopf, Edward Yang, Zachary DeVito, Martin Raison, Alykhan
  Tejani, Sasank Chilamkurthy, Benoit Steiner, Lu Fang, Junjie Bai, and Soumith
  Chintala.
\newblock Pytorch: An imperative style, high-performance deep learning library.
\newblock In {\em Advances in Neural Information Processing Systems 32}. 2019.

\bibitem{peng2018deepmimic}
Xue~Bin Peng, Pieter Abbeel, Sergey Levine, and Michiel Van~de Panne.
\newblock Deepmimic: Example-guided deep reinforcement learning of
  physics-based character skills.
\newblock {\em ACM Trans. Graph.}, 37(4), 2018.

\bibitem{peng2022ase}
Xue~Bin Peng, Yunrong Guo, Lina Halper, Sergey Levine, and Sanja Fidler.
\newblock Ase: Large-scale reusable adversarial skill embeddings for physically
  simulated characters.
\newblock {\em ACM Trans. Graph}, 41(4), 2022.

\bibitem{peng2021amp}
Xue~Bin Peng, Ze Ma, Pieter Abbeel, Sergey Levine, and Angjoo Kanazawa.
\newblock Amp: Adversarial motion priors for stylized physics-based character
  control.
\newblock {\em ACM Trans. Graph}, 40(4), 2021.

\bibitem{petrovich2021action}
Mathis Petrovich, Michael~J Black, and G{\"u}l Varol.
\newblock Action-conditioned 3d human motion synthesis with transformer vae.
\newblock In {\em ICCV}, 2021.

\bibitem{qin2022dexmv}
Yuzhe Qin, Yueh-Hua Wu, Shaowei Liu, Hanwen Jiang, Ruihan Yang, Yang Fu, and
  Xiaolong Wang.
\newblock Dexmv: Imitation learning for dexterous manipulation from human
  videos.
\newblock In {\em ECCV}, 2022.

\bibitem{ravi2020pytorch3d}
Nikhila Ravi, Jeremy Reizenstein, David Novotny, Taylor Gordon, Wan-Yen Lo,
  Justin Johnson, and Georgia Gkioxari.
\newblock Accelerating 3d deep learning with pytorch3d.
\newblock {\em arXiv:2007.08501}, 2020.

\bibitem{savva2016pigraphs}
Manolis Savva, Angel~X Chang, Pat Hanrahan, Matthew Fisher, and Matthias
  Nie{\ss}ner.
\newblock Pigraphs: learning interaction snapshots from observations.
\newblock {\em ACM Trans. Graph.}, 35(4), 2016.

\bibitem{schulman2017proximal}
John Schulman, Filip Wolski, Prafulla Dhariwal, Alec Radford, and Oleg Klimov.
\newblock Proximal policy optimization algorithms.
\newblock {\em arXiv preprint arXiv:1707.06347}, 2017.

\bibitem{shum2008interaction}
Hubert~PH Shum, Taku Komura, Masashi Shiraishi, and Shuntaro Yamazaki.
\newblock Interaction patches for multi-character animation.
\newblock {\em ACM Trans. Graph.}, 27(5), 2008.

\bibitem{starke2022deepphase}
Sebastian Starke, Ian Mason, and Taku Komura.
\newblock Deepphase: periodic autoencoders for learning motion phase manifolds.
\newblock {\em ACM Trans. Graph.}, 41(4), 2022.

\bibitem{starke2019neural}
Sebastian Starke, He Zhang, Taku Komura, and Jun Saito.
\newblock Neural state machine for character-scene interactions.
\newblock {\em ACM Trans. Graph.}, 38(6), 2019.

\bibitem{taheri2022goal}
Omid Taheri, Vasileios Choutas, Michael~J Black, and Dimitrios Tzionas.
\newblock Goal: Generating 4d whole-body motion for hand-object grasping.
\newblock In {\em CVPR}, 2022.

\bibitem{taheri2020grab}
Omid Taheri, Nima Ghorbani, Michael~J Black, and Dimitrios Tzionas.
\newblock Grab: A dataset of whole-body human grasping of objects.
\newblock In {\em ECCV}, 2020.

\bibitem{taylor2009factored}
Graham~W Taylor and Geoffrey~E Hinton.
\newblock Factored conditional restricted boltzmann machines for modeling
  motion style.
\newblock In {\em ICML}, 2009.

\bibitem{treuille2007near}
Adrien Treuille, Yongjoon Lee, and Zoran Popovi{\'c}.
\newblock Near-optimal character animation with continuous control.
\newblock In {\em ACM SIGGRAPH 2007 papers}. 2007.

\bibitem{villegas2017longterm}
Ruben Villegas, Jimei Yang, Yuliang Zou, Sungryull Sohn, Xunyu Lin, and Honglak
  Lee.
\newblock Learning to generate long-term future via hierarchical prediction.
\newblock In {\em ICML}, 2017.

\bibitem{wang2022towards}
Jingbo Wang, Yu Rong, Jingyuan Liu, Sijie Yan, Dahua Lin, and Bo Dai.
\newblock Towards diverse and natural scene-aware 3d human motion synthesis.
\newblock In {\em CVPR}, 2022.

\bibitem{wang2021synthesizing}
Jiashun Wang, Huazhe Xu, Jingwei Xu, Sifei Liu, and Xiaolong Wang.
\newblock Synthesizing long-term 3d human motion and interaction in 3d scenes.
\newblock In {\em CVPR}, 2021.

\bibitem{wang2020motion}
Jingbo Wang, Sijie Yan, Bo Dai, and Dahua Lin.
\newblock Scene-aware generative network for human motion synthesis.
\newblock In {\em CVPR}, 2021.

\bibitem{wang2022humanise}
Zan Wang, Yixin Chen, Tengyu Liu, Yixin Zhu, Wei Liang, and Siyuan Huang.
\newblock Humanise: Language-conditioned human motion generation in 3d scenes.
\newblock {\em NeurIPS}, 2022.

\bibitem{won2020scalable}
Jungdam Won, Deepak Gopinath, and Jessica Hodgins.
\newblock A scalable approach to control diverse behaviors for physically
  simulated characters.
\newblock {\em ACM Trans. Graph.}, 39(4), 2020.

\bibitem{wu2022saga}
Yan Wu, Jiahao Wang, Yan Zhang, Siwei Zhang, Otmar Hilliges, Fisher Yu, and
  Siyu Tang.
\newblock Saga: Stochastic whole-body grasping with contact.
\newblock In {\em ECCV}, 2022.

\bibitem{xiang2020sapien}
Fanbo Xiang, Yuzhe Qin, Kaichun Mo, Yikuan Xia, Hao Zhu, Fangchen Liu, Minghua
  Liu, Hanxiao Jiang, Yifu Yuan, He Wang, et~al.
\newblock Sapien: A simulated part-based interactive environment.
\newblock In {\em CVPR}, 2020.

\bibitem{xie2022chore}
Xianghui Xie, Bharat~Lal Bhatnagar, and Gerard Pons-Moll.
\newblock Chore: Contact, human and object reconstruction from a single rgb
  image.
\newblock In {\em ECCV}, 2022.

\bibitem{xu2021d3d}
Xiang Xu, Hanbyul Joo, Greg Mori, and Manolis Savva.
\newblock D3d-hoi: Dynamic 3d human-object interactions from videos.
\newblock {\em arXiv preprint arXiv:2108.08420}, 2021.

\bibitem{yang2022chopsticks}
Zeshi Yang, Kangkang Yin, and Libin Liu.
\newblock Learning to use chopsticks in diverse gripping styles.
\newblock {\em ACM Trans. Graph.}, 41(4), 2022.

\bibitem{zhang2021manipnet}
He Zhang, Yuting Ye, Takaaki Shiratori, and Taku Komura.
\newblock Manipnet: Neural manipulation synthesis with a hand-object spatial
  representation.
\newblock {\em ACM Trans. Graph.}, 40(4), 2021.

\bibitem{zhang2020phosa}
Jason~Y. Zhang, Sam Pepose, Hanbyul Joo, Deva Ramanan, Jitendra Malik, and
  Angjoo Kanazawa.
\newblock Perceiving 3d human-object spatial arrangements from a single image
  in the wild.
\newblock In {\em ECCV}, 2020.

\bibitem{zhang2020place}
Siwei Zhang, Yan Zhang, Qianli Ma, Michael~J Black, and Siyu Tang.
\newblock Place: Proximity learning of articulation and contact in 3d
  environments.
\newblock In {\em 3DV}, 2020.

\bibitem{zhang2022couch}
Xiaohan Zhang, Bharat~Lal Bhatnagar, Sebastian Starke, Vladimir Guzov, and
  Gerard Pons-Moll.
\newblock Couch: Towards controllable human-chair interactions.
\newblock In {\em ECCV}, 2022.

\bibitem{zhang2020generating}
Yan Zhang, Mohamed Hassan, Heiko Neumann, Michael~J Black, and Siyu Tang.
\newblock Generating 3d people in scenes without people.
\newblock In {\em CVPR}, 2020.

\bibitem{Zhao:ECCV:2022}
Kaifeng Zhao, Shaofei Wang, Yan Zhang, Thabo Beeler, , and Siyu Tang.
\newblock Compositional human-scene interaction synthesis with semantic
  control.
\newblock In {\em ECCV}, 2022.

\bibitem{zhou2022toch}
Keyang Zhou, Bharat~Lal Bhatnagar, Jan~Eric Lenssen, and Gerard Pons-Moll.
\newblock Toch: Spatio-temporal object-to-hand correspondence for motion
  refinement.
\newblock In {\em ECCV}, 2022.

\bibitem{zhou20196d}
Yi Zhou, Connelly Barnes, Lu Jingwan, Yang Jimei, and Li Hao.
\newblock On the continuity of rotation representations in neural networks.
\newblock In {\em CVPR}, 2019.

\end{thebibliography}
}

\end{document}